%% file: main.tex
\documentclass{article}
\input{paper_preamble.tex}
\input{paper_authors.tex}
\newcommand{\projecturl}{\url{https://longrca-bench.github.io/}}
\iclrfinalcopy
\begin{document}
\maketitle
\fancyhead{}
\input{paper_content.tex}
\end{document}

%% file: paper_preamble.tex
\usepackage{iclr2027_conference,times}
\usepackage[T1]{fontenc}
\input{math_commands.tex}

\usepackage{hyperref}
\hypersetup{hidelinks}
\usepackage{url}
\usepackage{graphicx}
\usepackage{booktabs}
\usepackage{tabularx}
\usepackage{array}
\usepackage{colortbl}
\usepackage{algorithm}
\usepackage{algpseudocode}
\usepackage[section]{placeins}
\usepackage{flafter}
\usepackage{wrapfig}
\usepackage{needspace}
\usepackage{float}
\usepackage{caption}
\usepackage{tcolorbox}
\definecolor{findingteal}{RGB}{0,83,86}
\newtcolorbox{findingbox}{colback=black!3,colframe=findingteal,
  boxrule=0.6pt,arc=2mm,left=5pt,right=5pt,top=3pt,bottom=3pt,
  before skip=4pt,after skip=4pt,fontupper=\normalsize}

\title{LongRCA Bench: Root-Cause Localization \\ 
in Long-Horizon Agent Trajectories}

%% file: math_commands.tex
\usepackage{amsmath,amsfonts,bm}

\def\eqref#1{equation~\ref{#1}}

\def\1{\bm{1}}

\DeclareMathAlphabet{\mathsfit}{\encodingdefault}{\sfdefault}{m}{sl}
\SetMathAlphabet{\mathsfit}{bold}{\encodingdefault}{\sfdefault}{bx}{n}



%% file: paper_authors.tex
\author{%
\begin{minipage}{\dimexpr\textwidth-2\tabcolsep\relax}
\centering
{\normalsize\bfseries
\mbox{Yunfei Zhang\textsuperscript{1,*}},
\mbox{Boyu Feng\textsuperscript{3,*}},
\mbox{Changhua Pei\textsuperscript{1,\textdagger}},
\mbox{Zexin Wang\textsuperscript{1}},
\mbox{Zhihuang Peng\textsuperscript{1}},
\mbox{Xinlong Liu\textsuperscript{2}},
\mbox{Hengyue Jiang\textsuperscript{2}},
\mbox{Difeng Ma\textsuperscript{1}},
\mbox{Jiayi Zhang\textsuperscript{1}},
\mbox{Yongzhou Yao\textsuperscript{4}},
\mbox{Yanan Zhao\textsuperscript{3}},
\mbox{Fei Sun\textsuperscript{4}},
\mbox{Yintong Huo\textsuperscript{5}},
\mbox{Zhaoyang Liu\textsuperscript{6}},
\mbox{Jingjing Li\textsuperscript{1}},
\mbox{Gaogang Xie\textsuperscript{1}},
\mbox{Dan Pei\textsuperscript{7}}\\[0.7em]
}
{\normalsize
\mbox{\textsuperscript{1}Computer Network Information Center, Chinese Academy of Sciences}\quad
\mbox{\textsuperscript{2}Hangzhou Institute for Advanced Study, University of Chinese Academy of Sciences}\quad
\mbox{\textsuperscript{3}Chongqing University}\quad
\mbox{\textsuperscript{4}Institute of Computing Technology, Chinese Academy of Sciences}\quad
\mbox{\textsuperscript{5}Singapore Management University}\quad
\mbox{\textsuperscript{6}Tongyi Lab, Alibaba Group}\quad
\mbox{\textsuperscript{7}Tsinghua University}\\[0.7em]
}
{\normalsize \textsuperscript{*}Equal contribution. \quad
\textsuperscript{\textdagger}Corresponding author.}
\end{minipage}
}

%% file: paper_content.tex
\let\maintextSavedFloatBarrier\FloatBarrier
\let\FloatBarrier\relax

\begin{abstract}
In long agent executions, an early error can persist through later actions
and checks, while evidence needed to trace its origin is dispersed across
the history. Short histories offer limited tests of recovering error
origins across substantial subsequent execution. We introduce
\textbf{LongRCA Bench}: 1,140 complete failed trajectories from five sources,
all human-annotated for responsible roles and earliest decisive root-cause
steps. Reference roots precede completion by a median of 48 recorded steps;
28.4\% of trajectories contain over 100 subsequent steps. With
DeepSeek-V4-Flash on the full benchmark, the strongest of five evaluated
baselines achieves 13.2\% exact root-step accuracy. We propose \textbf{Root-Cause
Trajectory Attribution (RCTA)}, a training-free method that organizes original
candidate records and explicit handoff instructions for attribution.
Segment summaries and a trajectory outline guide candidate retrieval;
available handoff records supply upstream instruction context for the final
instruction--execution comparison. With the same backbone and scoring protocol, RCTA reaches 24.1\%
exact root-step accuracy and 51.1\% responsible-role accuracy.
\textbf{LongRCA-Mini} provides 200 fixed trajectories for lower-cost
comparative screening. Even with RCTA, fewer than one quarter of reference
roots are recovered exactly.
\end{abstract}

\section{Introduction}

Failure attribution guides targeted debugging \citep{whowhen}. In long
executions, evidence spans plans, tool results, and verification
\citep{agentrx}. In Figure~\ref{fig:case0104}, repeated checks retain a nightly
lodging price as a two-night total despite correction of a separate
arithmetic error. Diagnosis requires recovering dispersed evidence
and separating error introduction from propagation or missed correction.

Existing benchmarks establish agent-, step-, and root-cause attribution
targets \citep{whowhen,whowhenpro,traceelephant,wang2026trajaudit}. Short
histories (Table~\ref{tab:benchmark_landscape}) offer limited tests of
recovering error origins across substantial subsequent activity, while
differing recording granularity prevents treating length alone as a controlled
difficulty measure.

We introduce \textbf{LongRCA Bench}\footnote{Project page and leaderboard:
\projecturl} for retrospective attribution in 1,140 complete, human-annotated
failed trajectories from five sources. The responsible workflow role and
root-cause step are predicted and scored independently. The root is the
earliest recorded step introducing a decisive error relevant to the failure
and unrepaired at completion. Reference roots precede completion by a median
of 48 recorded steps; 28.4\% of trajectories contain more than 100 subsequent
steps.

\begingroup
\setlength{\columnsep}{12pt}
\begin{wrapfigure}{R}{0.48\textwidth}
\vspace{-5pt}
\centering
\includegraphics[width=\linewidth]{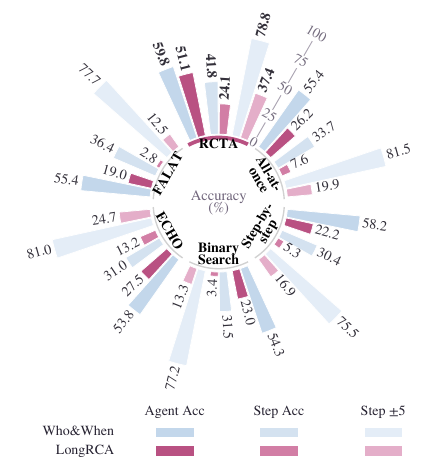}
\captionsetup{font=small,hypcap=false,justification=raggedright,singlelinecheck=false}
\caption{All six evaluated methods score lower on LongRCA across Agent Acc,
Step Acc, and Step $\pm5$. Accuracy (\%) uses DS-V4-Flash on the complete
Who\&When ($n=184$) and LongRCA ($n=1{,}140$) sets, with each trajectory
equally weighted within its dataset.}
\label{fig:whowhen_comparison}
\vspace{-5pt}
\end{wrapfigure}
\raggedright
\rightskip=0pt plus 2em\relax
\hyphenpenalty=10000
\exhyphenpenalty=10000
As shown in Figure~\ref{fig:whowhen_comparison}, the five evaluated
baselines achieve 30.4--36.4\% exact root-step accuracy on
Who\&When~\citep{whowhen} but only 2.8--13.2\% on LongRCA
using DeepSeek-V4-Flash (DS-V4-Flash). The datasets
also differ in domains and workflows, so this gap does not isolate a
trajectory-length effect.

Locating a failure-relevant action does not establish whether it introduced
the error or inherited it from an earlier instruction.
We propose
\textbf{Root-Cause Trajectory Attribution (RCTA)}, a training-free method
that first retrieves original candidate evidence from long logs, then
compares it with available recorded upstream instructions to distinguish
inherited errors from errors newly introduced during execution. This
separates locating failure-relevant activity from identifying where the
decisive error originated.

With the same DS-V4-Flash backbone and scoring protocol, RCTA raises exact
root-step accuracy from the strongest evaluated baseline's 13.2\% to
24.1\% (Table~\ref{tab:main_results}). Nevertheless, fewer than one quarter
of reference roots are recovered exactly.
\par\noindent\textbullet\enspace \textbf{A benchmark for tracing error origins.} LongRCA Bench provides
1,140 human-annotated failed trajectories from five sources, accompanied by
\textbf{LongRCA-Mini}, a fixed 200-trajectory subset for lower-cost preliminary
comparisons. Headline results and leaderboard evaluation use the full benchmark.

\par\noindent\textbullet\enspace \textbf{A method for tracing errors to their source.} RCTA retrieves
relevant original records from long logs. It then checks available earlier
instructions to judge whether an error was already present in those
instructions or was introduced during execution.

\par\noindent\textbullet\enspace \textbf{A comparison of six attribution methods.} We compare six
methods on Full and across five backbones on Mini, and test RCTA with
individual components removed. The results show where localization remains
difficult and how individual components affect accuracy, the evidence
retained, and output validity.

\par
\WFclear
\endgroup

\section{Related Work}

\begin{table}[!t]
\centering
\caption{
Comparison of resources for agent failure analysis.
LongRCA supports responsible-role and earliest-root-step attribution
over long execution trajectories collected without error injection
and fully reviewed by humans.
}
\label{tab:benchmark_landscape}

\begingroup
\small
\setlength{\tabcolsep}{3pt}
\renewcommand{\arraystretch}{1.18}
\renewcommand{\tabularxcolumn}[1]{m{#1}}

\begin{tabularx}{\linewidth}{
    @{}
    >{\raggedright\arraybackslash}m{0.370\linewidth}
    >{\raggedright\arraybackslash}m{0.105\linewidth}
    >{\raggedright\arraybackslash}X
    >{\raggedleft\arraybackslash}m{0.160\linewidth}
    >{\raggedleft\arraybackslash}m{0.125\linewidth}
    @{}
}
\toprule
\textbf{Benchmark}
&
\shortstack[l]{\textbf{Error}\\\textbf{Source}}
&
\shortstack[l]{\textbf{Diagnostic}\\\textbf{Targets}}
&
\shortstack[r]{\textbf{Trajectories}\\Total (Reviewed)}
&
\shortstack[r]{\textbf{Avg. length}\\(steps)}
\\
\midrule

RootSE~\citep{wang2026trajaudit}
& Natural
& Step\textsuperscript{a}
& 102 (102)
& 51.7
\\
\addlinespace[2pt]

HORIZON~\citep{horizon}
& Natural
& Type
& 1995 (40)
& --
\\
\addlinespace[2pt]

TraceElephant~\citep{traceelephant}
& Natural
& Component, Step\textsuperscript{b}
& 220 (220)
& --
\\
\addlinespace[2pt]

Failure as a Process~\citep{failureprocess}
& Natural
& Steps\textsuperscript{c}, Type
& 1184 (1184)
& 42.0
\\
\addlinespace[2pt]

Who\&When~\citep{whowhen}
& Natural
& Agent, Step\textsuperscript{a}
& 184 (184)
& 22.2
\\
\addlinespace[2pt]

Who\&When Pro~\citep{whowhenpro}
& Synthetic
& Agent, Step\textsuperscript{a}, Type
& 12,326 (100)
& 7.5
\\

\midrule
\textbf{LongRCA Bench (ours)}
& \textbf{Natural}
& \textbf{Agent, Step\textsuperscript{a}}
& \textbf{1140 (1140)}
& \textbf{156.3}
\\
\bottomrule
\end{tabularx}

\par\vspace{4pt}
\parbox{\linewidth}{%
    \footnotesize
    \textsuperscript{a}Earliest decisive error,
    \textsuperscript{b}Earliest point at which failure becomes inevitable,
    \textsuperscript{c}Three markers: decisive error, empirical lock-in,
    and first observable sign of error.
    Annotation protocols and step granularity differ across benchmarks, so
    mean lengths are not directly comparable measures of difficulty;
    -- denotes an unreported mean. Appendix~\ref{app:related_work} details
    categories and counting conventions.
}
\endgroup
\end{table}

Who\&When formulates attribution as identifying a responsible agent and a
decisive step \citep{whowhen}. AgentRx targets the earliest critical failure
from which an agent does not recover, while RootSE, introduced with TrajAudit,
annotates earliest decisive errors in repository-level coding traces
\citep{agentrx,wang2026trajaudit}. Failure as a Process separately marks
decisive errors, empirical unrecoverability, and first observable signs
\citep{failureprocess}. Table~\ref{tab:benchmark_landscape} compares these
resources. HORIZON varies intrinsic
task horizon under controlled conditions to study performance and failure
patterns \citep{horizon}. This horizon is distinct from recorded trajectory
length. LongRCA Bench evaluates
attribution in complete failed executions with long recorded histories,
especially when substantial activity follows the reference root.

Attribution methods differ in how they organize evidence and trace error
origins. ECHO combines hierarchical context with analyst consensus
\citep{echo}, while SAFARI investigates candidate faults through active
trace reading and search with persistent short-term memory \citep{safari}.
FALAT uses typed dependencies among decisions, tool outputs, and messages
to separate error sources from downstream carriers and guide candidate
pruning and verification \citep{falat}. CHIEF combines hierarchical
backtracking with local, planning-control, and data-flow attribution,
screening out subsequently repaired deviations \citep{chief}.

RCTA organizes attribution around retrieved original candidate records.
After candidate retrieval, it adds available explicit handoffs as upstream
instruction evidence, retaining the original text and step IDs. The final
comparison examines whether execution inherited an erroneous instruction
or introduced a new decisive error under the task requirements. This
augmentation depends on logged handoff instructions; when none are
recorded, attribution proceeds from the retrieved candidate evidence
(Section~\ref{sec:method}).
Appendix~\ref{app:related_work} extends the discussion of diagnostic resources
and mechanisms and details the table's comparison conventions.

\section{LongRCA Bench}
\label{sec:benchmark}

\subsection{Task definition}
\label{sec:task_definition}
\label{sec:illustrative_case}

We study retrospective failure attribution over completed failed executions,
using the full history to distinguish error introduction from subsequent
execution. This retrospective setting uses the completed record rather than
a trajectory prefix for online failure detection. A trajectory
$H=(h_0,\ldots,h_{T-1})$ has stable 0-based indices
and recorded source roles. The responsible role is the recorded workflow role
judged responsible for task failure under the annotation protocol. Methods
return a role $\hat{\rho}$ and root step
$\hat{r}\in\{0,\ldots,T-1\}$ in separate fields, scored independently against
$(\rho^*,r^*)$. The reference role must belong to the recorded role set but
need not emit the reference root step. The role prediction space is likewise
limited to recorded workflow roles. Predicted roles are read directly,
never inferred from the predicted step. Each reference includes a
trajectory-grounded human rationale for evidence review, which is not scored.

The reference root $r^*$ is the earliest recorded step introducing a decisive
error that is relevant to the failed execution and remains unrepaired at completion.
Repaired errors and later steps that merely execute, propagate, or expose an
existing error are excluded. Under a handoff, the instruction step is selected
if it already contains the decisive error and the recipient carries it out
without repair. A later recipient step qualifies only if it introduces the
decisive error through a deviation or an error absent from the instruction.
Appendix~\ref{app:rules} specifies the complete boundary rules.

\begin{figure}[!t]
\centering
\includegraphics[width=\linewidth]{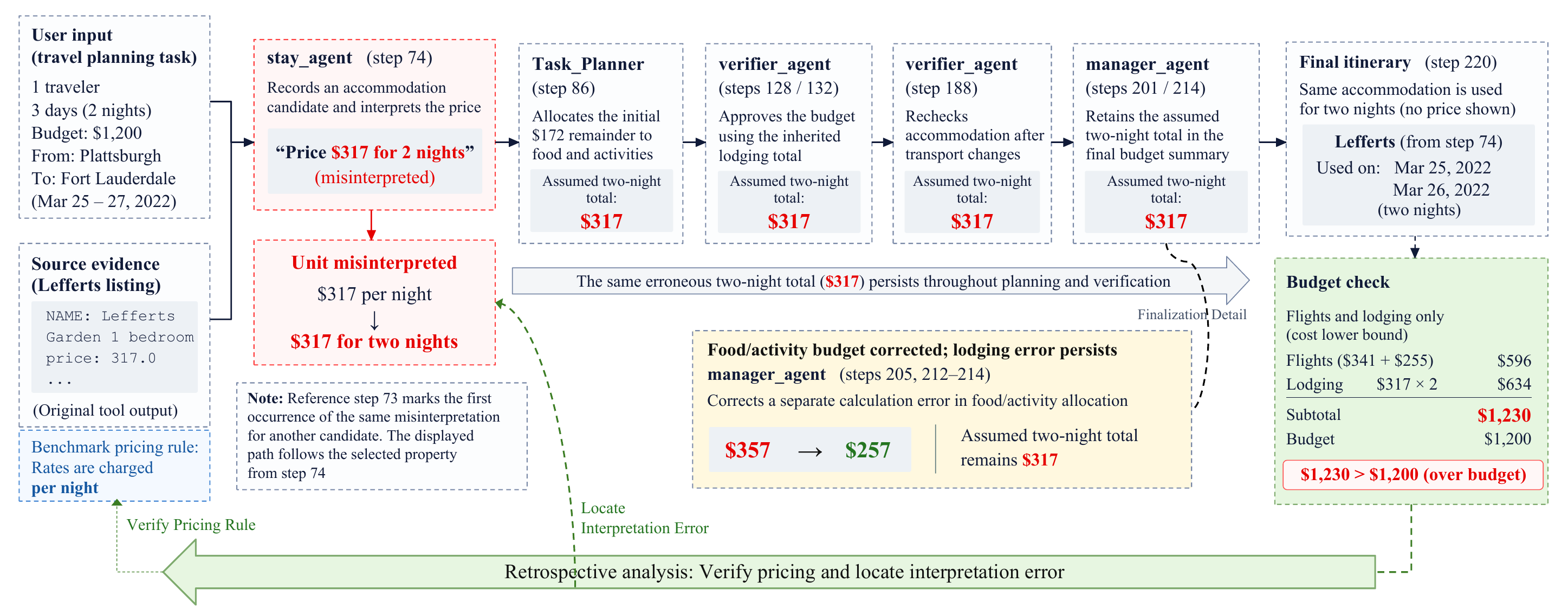}
\caption{A price-unit misinterpretation persists through planning and
verification despite a local budget correction. The human reference identifies
\texttt{stay\_agent} and step 73 within a single output containing candidate
records at steps 73--77. The displayed path follows the ultimately selected
property from step 74. Green arrows indicate retrospective diagnosis using
the final itinerary and TravelPlanner's nightly pricing rule.}
\label{fig:case0104}
\end{figure}

Figure~\ref{fig:case0104} illustrates the persistence of a lodging-price
misinterpretation. The human reference assigns responsibility to
\texttt{stay\_agent} and marks step 73 within a single output whose
candidate-recording calls, steps 73--77, treat nightly prices as two-night
totals. The figure follows the ultimately selected property from step 74
through subsequent planning and verification. A later correction to the
food/activity budget leaves its lodging total unchanged. The case illustrates
why attribution requires tracing early erroneous records through their
subsequent use and distinguishing persistent errors from local repairs.
Appendix~\ref{app:case_study} details the reference annotation, the
selected-property path, and the budget diagnosis.

\subsection{Source curation and trajectory normalization}

We retain evaluator-confirmed failed task attempts from SWE-bench Pro (SWE, 128),
Terminal Bench 2 (TB2, 42), TravelPlanner (TP, 685), VitaBench (Vita, 108),
and WebArena Verified (WA, 177)
\citep{swebenchpro,terminalbench2,travelplanner,vitabench,webarena_verified}.
These sources span different task domains and workflows, including fixed-role
teams, group-chat coordination, and sequential execution. Failures arise from
ordinary execution rather than injected errors. Infrastructure, smoke-test,
and debug runs are excluded because they are not completed task attempts.
Appendix~\ref{app:source_provenance} documents source configurations and
trajectory generators.

Normalization preserves original content, roles, and temporal order with
stable 0-based indices, without synthesizing intermediate actions or
artificially subdividing source events. The shared interface does not imply
identical event granularity. Methods receive only \texttt{question\_ID} and
\texttt{history}, including the task instruction; reference role, step, and
rationale are withheld. Source evaluator outcomes guide selection and
annotation, while feedback already recorded within history is retained
(Appendix~\ref{app:schema}).

\subsection{Human annotation and quality control}

Twenty-two master's and doctoral students in computer science annotated the
trajectories. Calibration on 100 trajectories refined history navigation,
root-cause search, and evidence recording, while the root definition and
decision rules remained unchanged for full-scale annotation. The same
annotator pool participated in both stages. Annotators independently reviewed the task,
complete history, and source evaluator outcome to submit a responsible role,
root step, and supporting rationale. DeepSeek V4 Flash Preview supplied
factual activity summaries for navigation without root-cause judgments.
Reference labels and rationales required original trajectory evidence.

Full-scale annotation provides 1,343 completed submissions with verifiable
role, step, and rationale fields over 1,041 trajectories. The 294 multiply
annotated trajectories yield 310 pairs: role agreement is 66.1\%, step exact
agreement is 40.3\%, and $\pm5$ agreement is 50.3\%; 38.1\% of pairs differ
by more than 20 steps. With DS-V4-Flash, RCTA ranks first in Step Acc
in all 10,000 draws of existing human labels on the 294-trajectory subset
(Appendix~\ref{app:annotation}).

Where multiple valid annotations were available, labels were compared field
by field and exact matches retained as consensus. Two experts reviewed
disagreements against original tasks, histories, evaluator outcomes, and
rationales. Final quality control re-examined all retained trajectories,
including calibration cases, and removed cases with unresolved ambiguity,
invalid labels, or inadequate support. Of 1,200 reviewed trajectories, 60
were excluded, leaving 1,140 human-annotated failed executions. Each final
record retains independent reference fields for role and step, together with
a supporting rationale.

\subsection{Benchmark characteristics and LongRCA-Mini}
\label{sec:characteristics}
\label{sec:mini_construction}

\begin{figure}[!htbp]
\centering
\includegraphics[width=0.90\linewidth]{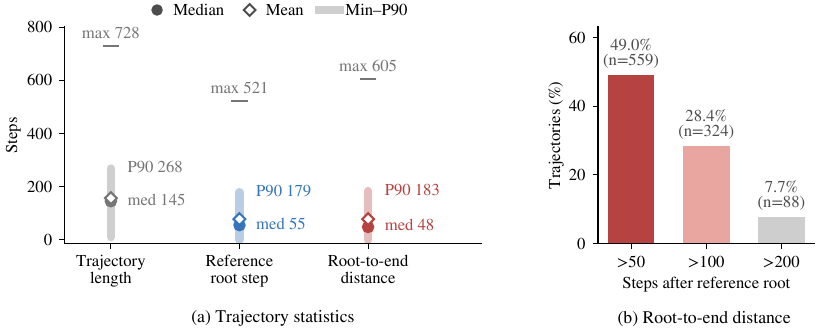}
\caption{Long-horizon characteristics ($n=1,140$). (a) Medians (circles),
means (diamonds), minimum--P90 ranges, and maxima of trajectory length,
reference root step, and root-to-end distance. (b) Fractions with more than
50, 100, or 200 recorded steps after the reference root.}
\label{fig:delay}
\end{figure}

Trajectory length counts recorded steps in a completed execution
(mean 156.3; median 145). Root-to-end distance, $T-1-r^*$, counts steps
after the reference root (median 48); 28.4\% of trajectories have more
than 100 such steps (Figure~\ref{fig:delay}). These Full summaries retain
natural source proportions. This distance measures subsequent activity,
not error propagation or time to observable failure. Additional statistics
and length-group accuracy appear in Appendix~\ref{app:benchmark_statistics}
and Section~\ref{sec:length_results}, respectively.

LongRCA-Mini supports preliminary comparisons with 200 fixed trajectories,
sampled uniformly without replacement using 40 per source. Sample size is
calibrated using Full predictions; individual trajectories are selected
without predictions or reference labels, and observed results do not trigger
resampling.

Fidelity is assessed against Full Macro-5, which weights the five sources
equally (Section~\ref{sec:settings}). Across six methods and three accuracy
metrics under DS-V4-Flash, the fixed subset has mean and maximum absolute
score deviations of 1.665 and 3.771 pp, respectively, using cached Full
predictions. These values quantify subset-induced differences under the
evaluated configuration. Appendix~\ref{app:mini} provides the calibration
protocol and detailed fidelity results.

\maintextSavedFloatBarrier

\section{Root-Cause Trajectory Attribution}
\label{sec:method}

\begin{figure}[!t]
\centering
\includegraphics[width=\linewidth]{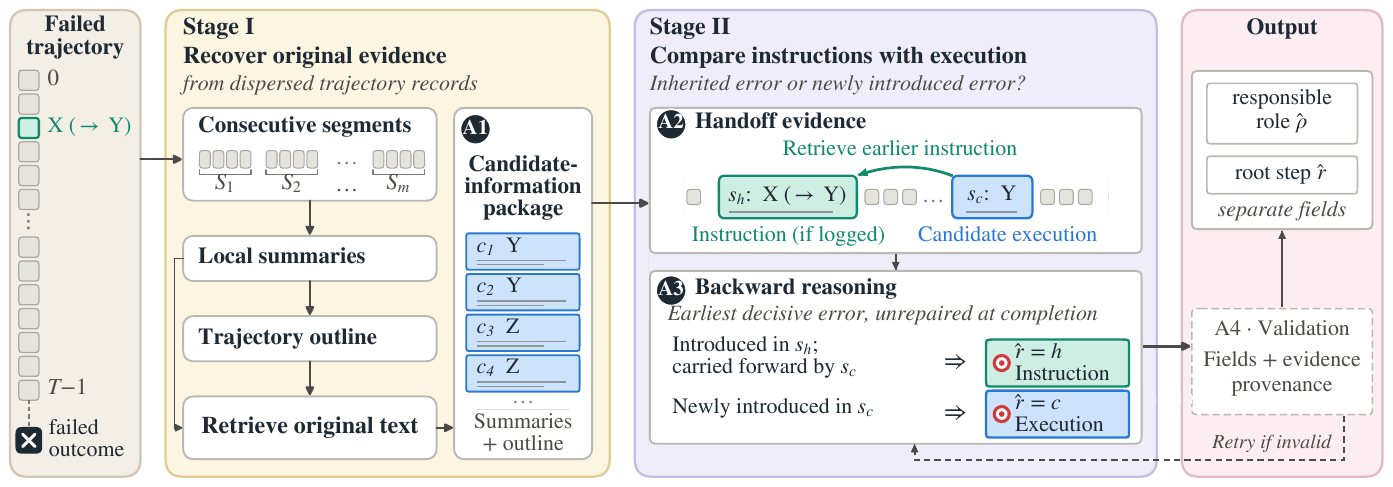}
\caption{RCTA has two stages: Stage I constructs the candidate-information
package (A1); Stage II retrieves available handoff evidence (A2) and performs
backward instruction--execution reasoning (A3). Output and evidence validation
(A4) checks the resulting attribution output and its evidence provenance.}
\label{fig:pipeline}
\end{figure}

RCTA organizes original candidate records and available handoff instructions
for attribution (Figure~\ref{fig:pipeline}). Candidate
retrieval and handoff augmentation determine which evidence is presented to
the attribution stage; an auxiliary layer validates output fields and cited
evidence.

\subsection{Stage I: Failure-relevant candidate recall}

Stage I builds the candidate-information package (A1). A \emph{segment}
contains consecutive recorded steps with original IDs. Task-conditioned
local summaries propose failure-relevant steps, using available preceding
overlap records as context. A trajectory outline aggregates subgoals and
conflicts across summaries. RCTA pools and prioritizes proposed step IDs,
then retrieves their original text to form the base-candidate set before
handoff augmentation. A1 combines these records, local summaries, and the
outline, allowing attribution to inspect source evidence alongside the
context guiding recall. Appendix~\ref{app:prompts} specifies segmentation,
retrieval rules, and prompts; Algorithm~\ref{alg:rcta_inference} gives the
inference sequence.

\subsection{Stage II: Handoff-aware backward attribution}

A \emph{handoff instruction} is a logged message in which a sender instructs
a recipient. Handoff evidence retrieval (A2) augments the base candidates with
such earlier records. For executor or verifier candidates, RCTA retrieves the
nearest preceding handoff whose recipient exactly matches the candidate's
recorded name (the role classes and matching rules are in
Appendix~\ref{app:prompts}). For
other candidates, it retrieves the nearest earlier handoff as plan context.
Candidate and handoff records retain their original text and step IDs.
Instructions are attached only when present; missing instructions are not
synthesized.

Backward instruction--execution reasoning (A3) evaluates candidate records
against task requirements and, when available, preceding handoff instructions.
It judges whether the decisive error was already present upstream or newly
introduced during execution.
Following Section~\ref{sec:task_definition}, the decision targets the earliest
supported introduction of a failure-relevant decisive error that remains
unrepaired at completion. An erroneous instruction faithfully executed without
repair can therefore be selected instead of its downstream candidate.
Retrieved handoffs can supply root steps outside
the base-candidate set. Original texts support this judgment, while summaries
and the outline provide context. Without handoffs, attribution proceeds from
the existing evidence.

The final attribution call returns separate responsible-role and root-step
fields. Validation (A4) checks field validity and the provenance of cited
handoff evidence, retaining valid fields independently.
Appendix~\ref{app:prompts} details retries and output handling.

\maintextSavedFloatBarrier

\begingroup
\raggedbottom
\setlength{\textfloatsep}{12pt plus 2pt minus 2pt}
\setlength{\intextsep}{6pt plus 2pt minus 2pt}
\captionsetup{skip=6pt}
\section{Evaluation}
\label{sec:experiments}

\subsection{Experimental settings}
\label{sec:settings}

We examine how attribution methods perform across sources and log
characteristics (RQ1), how their performance varies across backbones and
metrics (RQ2), and how RCTA component configurations relate to evidence
retention, output validity, and localization accuracy (RQ3).

The full-benchmark comparison evaluates six training-free methods on all 1,140 failed
trajectories using DeepSeek-V4-Flash (DS-V4-Flash). Alongside RCTA,
All-at-once reads the complete log in one attribution prompt, Step-by-step
scans it sequentially, and Binary search recursively bisects it
\citep{whowhen}. ECHO uses hierarchical context and consensus voting
\citep{echo}; FALAT traces dependencies among decisions, tool outputs, and
messages to localize error-introducing steps \citep{falat}. All methods use the same LongRCA Bench
attribution targets and scoring rules.

Agent Acc compares the explicit responsible-role output with its reference.
Step Acc requires an exact match to the 0-based reference root step; Step
$\pm5$ permits absolute step error at most five. Role and step are scored
independently, with invalid fields counted as errors for the corresponding
targets. Missing or unrecognized roles are invalid; missing, nonnumeric, or
out-of-range steps fail both step metrics. All expected trajectories remain
in accuracy denominators. Full
weights trajectories equally, preserving source proportions. Mini uses
$S_{\mathrm{Macro\text{-}5}}=\frac{1}{5}\sum_{b=1}^{5}S_b$, where $S_b$ is the
current metric's source score. Both sides of Full--Mini fidelity use Macro-5.
With equal source quotas, pooled accuracy equals Macro-5; this equality
need not hold for MAE when valid-output counts differ.

Mini experiments share the fixed sample; fidelity reuses cached Full predictions.
Ablations share segmentation and
summary caches across conditions. Appendix~\ref{app:scoring_protocols}
provides normalization, MAE, and protocol details.

\subsection{Full-benchmark attribution results (RQ1)}
\label{sec:full_results}

\begin{table}[!ht]
\centering
\caption{Full-benchmark accuracies (\%) using DS-V4-Flash. Overall weights
trajectories equally; per-source columns report Step Acc. Bold marks the
highest observed value in each column.}
\label{tab:main_results}
\normalsize
\fontencoding{OT1}\selectfont
\setlength{\tabcolsep}{3pt}
\renewcommand{\arraystretch}{1.10}
\begin{tabularx}{\linewidth}{@{}>{\raggedright\arraybackslash}p{1.12in}*{8}{>{\raggedleft\arraybackslash}X}@{}}
\toprule
\textbf{Method} & \multicolumn{3}{c}{\textbf{Overall}} & \multicolumn{5}{c}{\textbf{Per-source Step Acc}} \\
\cmidrule(lr){2-4}\cmidrule(lr){5-9}
 & \shortstack{Agent\\Acc} & \shortstack{Step\\Acc} & \shortstack{Step\\$\pm5$} & SWE & TB2 & TP & Vita & WA \\
\midrule
All-at-once & 26.2 & 7.6 & 19.9 & 1.6 & 7.1 & 4.5 & 19.4 & 16.9 \\
Step-by-step & 22.2 & 5.3 & 16.9 & 0.8 & 4.8 & 3.5 & 12.0 & 11.3 \\
Binary search & 23.0 & 3.4 & 13.3 & 0.8 & 4.8 & 2.0 & 12.0 & 5.1 \\
ECHO & 27.5 & 13.2 & 24.7 & 7.8 & 21.4 & 8.9 & 22.2 & 26.0 \\
FALAT & 19.0 & 2.8 & 12.5 & 2.3 & 2.4 & 2.9 & 3.7 & 2.3 \\
\textbf{RCTA} & \textbf{51.1} & \textbf{24.1} & \textbf{37.4} & \textbf{38.3} & \textbf{26.2} & \textbf{18.8} & \textbf{25.0} & \textbf{33.3} \\
\bottomrule
\end{tabularx}
\end{table}

On Full with DS-V4-Flash, RCTA has the highest observed values on all three
accuracy metrics, but exact root recovery remains below one quarter
(Table~\ref{tab:main_results}).
Role accuracy also requires comparison with source-specific priors: always
predicting each source's most frequent reference role yields 40.3\% Agent
Acc, 10.9 percentage points below RCTA (Table~\ref{tab:role_priors}).
This baseline uses role frequencies from the evaluated set. Identifying
the responsible role and locating the originating error remain distinct
targets; success on the former does not establish success on the latter.

RCTA's Step Acc lead appears in every source, so its overall advantage is
not confined to TravelPlanner, the largest subset. The between-source
comparisons remain descriptive because the represented tasks and workflows
differ. Appendix~\ref{app:full_metrics} retains the complete overall
metrics, including Root MAE; Appendix~\ref{app:stratified} supplies the
source visualization and stratified numeric results.

The cross-dataset comparison in Figure~\ref{fig:whowhen_comparison} uses
the same DS-V4-Flash backbone; Appendix~\ref{app:whowhen} provides the
protocol, results, and metric interpretation.

\subsection{Localization across log characteristics (RQ1)}
\label{sec:length_results}

\begin{figure}[!htbp]
\centering
\includegraphics[width=\linewidth]{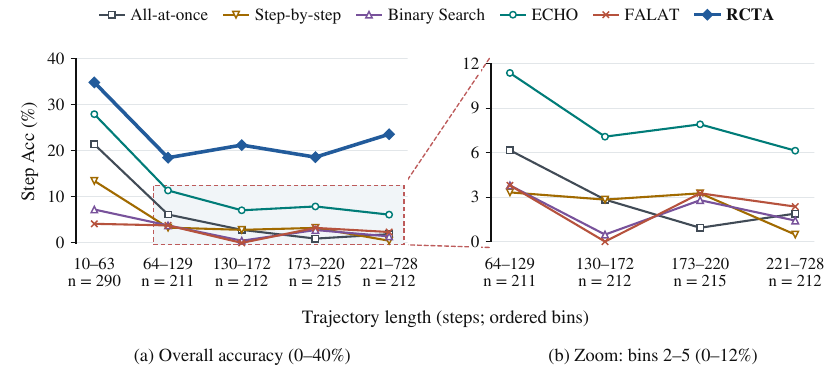}
\caption{Step Acc on all 1,140 trajectories with DS-V4-Flash. Methods share
length-only groups without splitting equal lengths. Panel (a) shows all groups;
panel (b) magnifies the 0--12\% region of groups 2--5. Unequal-width groups are
equally spaced; connecting lines are visual guides.}
\label{fig:length_accuracy}
\end{figure}

All methods share groups defined only by recorded length, without splitting
equal lengths (Figure~\ref{fig:length_accuracy}).
RCTA has the highest observed Step Acc in every group, yet all six methods
score lower in each longer group than in the shortest group.
Variation among the longer groups is non-monotonic: RCTA rises from
18.6\% to 23.6\% between the final two groups.

The longest group mainly
contains 221--400-step trajectories (170 of 212), so its score does not
characterize only the extreme longest runs. Source and workflow composition
also differ across groups, preventing interpretation as an independent
causal effect of length. Appendix~\ref{app:stratified} gives the numeric
results, partition algorithm, and four- and six-group sensitivity checks.

Within-source analyses further examine root-to-end distance and log text
volume. On TravelPlanner and VitaBench, all six methods have lower Step Acc
and Step $\pm5$ in the highest-character group than in the lowest-character
group. Patterns vary across sources, and localization performance does not
consistently decline with root-to-end distance
(Appendix~\ref{app:within_source}).

\subsection{Cross-backbone comparison on LongRCA-Mini (RQ2)}
\label{sec:mini_backbones}

\begin{table}[!htbp]
\centering
\caption{Agent and Step Acc (\%, Macro-5) on the fixed 200-trajectory Mini
(40 per source). V4 denotes DeepSeek-V4. Bold marks the highest point
estimate per backbone and metric; complete metrics and paired analyses are
in Appendix~\ref{app:mini_backbones}.}
\label{tab:mini_backbones}
\small
\fontencoding{OT1}\selectfont
\setlength{\tabcolsep}{2pt}
\renewcommand{\arraystretch}{1.08}
\begin{tabularx}{\linewidth}{@{}>{\raggedright\arraybackslash}p{0.96in}*{10}{>{\raggedleft\arraybackslash}X}@{}}
\toprule
\textbf{Method} & \multicolumn{2}{c}{\textbf{V4-Flash}} &
\multicolumn{2}{c}{\textbf{V4-Pro}} &
\multicolumn{2}{c}{\textbf{GPT-5.6 Sol}} &
\multicolumn{2}{c}{\textbf{Kimi K3}} &
\multicolumn{2}{c}{\shortstack{\textbf{Qwen3.5}\\\textbf{35B-A3B}}} \\
\cmidrule(lr){2-3}\cmidrule(lr){4-5}\cmidrule(lr){6-7}\cmidrule(lr){8-9}\cmidrule(lr){10-11}
 & Agent & Step & Agent & Step & Agent & Step & Agent & Step & Agent & Step \\
\midrule
All-at-once & 42.5 & 7.5 & 50.0 & 11.5 & 53.0 & 17.0 & 54.5 & 11.0 & \textbf{51.0} & 3.5 \\
Step-by-step & 38.5 & 6.5 & 47.5 & 10.5 & 49.0 & 13.0 & 44.0 & 9.5 & 46.5 & 4.5 \\
Binary search & 40.5 & 7.5 & 45.5 & 10.0 & 51.5 & 15.5 & 42.0 & 12.0 & 40.0 & 5.0 \\
ECHO & 45.0 & 13.0 & 53.5 & 18.0 & 55.5 & 18.5 & 55.0 & 20.5 & 48.5 & 10.0 \\
FALAT & 35.5 & 2.5 & 48.0 & 14.5 & 52.0 & 14.0 & 52.0 & 18.0 & 41.0 & 11.0 \\
\textbf{RCTA} & \textbf{60.5} & \textbf{27.0} & \textbf{61.0} & \textbf{28.0} & \textbf{64.5} & \textbf{30.0} & \textbf{61.5} & \textbf{31.5} & 48.5 & \textbf{13.5} \\
\bottomrule
\end{tabularx}
\end{table}

On LongRCA-Mini, the six-method comparison spans five backbones
(Table~\ref{tab:mini_backbones}). RCTA leads Agent Acc on four.
Qwen is the exception: All-at-once leads Agent Acc and ECHO leads Step
$\pm5$.

The comparisons do not establish significant superiority on every backbone
or isolate effects of model scale. Appendix~\ref{app:mini_backbones} retains all metrics, paired
tests, and their endpoint-family sensitivity. Reference-label overlap also
shows complementary exact localizations between RCTA and the five-baseline
union on Mini (Appendix~\ref{app:mini_backbones}).
Across the five evaluated backbones, RCTA has the highest Step Acc point
estimate, but does not lead every accuracy metric.

\maintextSavedFloatBarrier

\subsection{Component analysis (RQ3)}
\label{sec:ablation}

\begin{table}[!ht]
\centering
\caption{Paired ablations on the fixed Mini sample (Macro-5 percentages).
Cov. measures valid-output coverage;
$p_{\mathrm{Holm}}$ adjusts the four Step Acc tests.}
\label{tab:ablation}
\normalsize
\fontencoding{OT1}\selectfont
\setlength{\tabcolsep}{2.2pt}
\renewcommand{\arraystretch}{1.10}
\begin{tabularx}{\linewidth}{@{}>{\raggedright\arraybackslash}Xrrrrr@{}}
\toprule
\textbf{Setting} & \shortstack{Agent\\Acc$\uparrow$} &
\shortstack{Step\\Acc$\uparrow$} & \shortstack{Step\\$\pm5\,\uparrow$} &
Cov.$\uparrow$ & $p_{\mathrm{Holm}}$ \\
\midrule
\textbf{RCTA (all components)} & 56.0 & 28.0 & 41.0 & 97.0 & -- \\
\hspace*{0.8em}w/o candidate-information package (A1) & 53.5 & 22.0 & 39.0 & 91.5 & .17 \\
\hspace*{0.8em}w/o handoff evidence (A2) & 53.5 & 24.5 & 38.0 & 97.0 & .75 \\
\hspace*{0.8em}w/o backward attribution (A3) & 51.0 & 25.0 & 41.0 & 97.5 & .75 \\
\hspace*{0.8em}w/o validation (A4) & 54.0 & 26.5 & 39.5 & 89.0 & .75 \\
\bottomrule
\end{tabularx}
\end{table}

This study uses dedicated ablation inference settings, distinct from the
Full-benchmark configuration (Appendix~\ref{app:prompts}). All conditions
share trajectories, reference labels, segmentation, and summary caches.
The A1 control uses equally many uniformly positioned
excerpts and masks location and provenance cues, testing the complete
candidate-information package. Removing A2 omits handoff evidence; removing
A3 retains the original excerpts but removes the backward-attribution
package; removing A4 omits validation-driven retry and output sanitization.
Appendices~\ref{app:prompts} and~\ref{app:scoring_protocols} specify these
controls and scoring rules.

All four removals lower Step Acc point estimates, but none passes Holm
correction across the four comparisons ($p_{\mathrm{Holm}}\geq.17$;
Table~\ref{tab:ablation}). Non-significance does not establish equivalence;
Appendix~\ref{app:component_diagnostics} reports the paired intervals and
complete results.

\begin{figure}[!ht]
\centering
\includegraphics[width=0.90\linewidth]{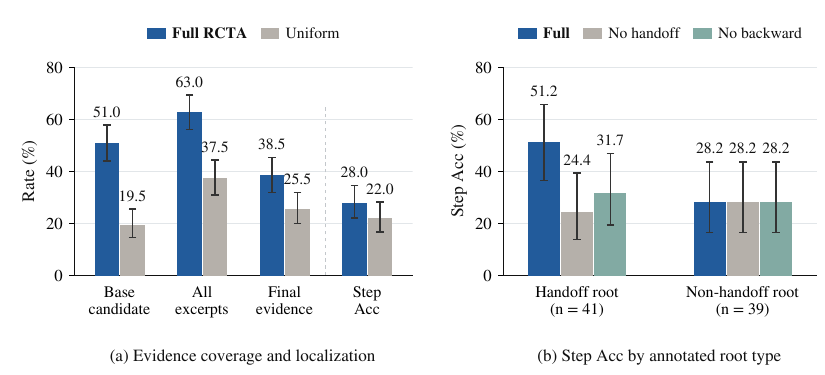}
\caption{Component diagnostics. (a) Reference-root hit rates and final Step
Acc (all $n=200$). (b) Step Acc on handoff-containing trajectories by
reference root: handoff ($n=41$) or non-handoff ($n=39$). Full and A3 share
the original excerpts. Error bars: 95\% Wilson intervals; subgroup
contrasts are exploratory.}
\label{fig:ablation_mechanism}
\end{figure}


The candidate-information package shows a larger gain in reference-root
retention than in final Step Acc (Figure~\ref{fig:ablation_mechanism}(a));
having the root in the evidence does not ensure correct localization.
Within the 80 component-study trajectories containing explicit handoffs,
the observed Step Acc differences for A2 and A3 are confined to the
handoff-root subgroup (panel (b)). These
exploratory comparisons cover the shared diagnosis--execution--verification
workflow of SWE-bench Pro and Terminal-Bench 2; they do not evaluate
trajectories without logged handoffs. Validation's clearest observed change
is valid-output coverage, from 89\% to 97\%, with exact-root correctness
changing on three of 200 trajectories.

\section{Discussion and Limitations}
\label{sec:discussion}

Our evaluation addresses retrospective attribution from complete failed
trajectories. A reference root need not be recognizable when it occurs;
online detection requires evaluation using only the observed execution
prefix, including warning timeliness and false-alarm rates. Whether warnings
or interventions improve task outcomes remains untested. Evidence validation
checks output fields and cited provenance, without establishing causal
correctness. Agreement with human role and step labels likewise does not
evaluate rationale quality or complete causal-chain reconstruction.
Generalization is limited to the represented task domains, workflows, and
inference settings.
The comparisons do not establish an advantage under matched token or cost
budgets (Appendix~\ref{app:prompts}).

\section{Conclusion}
\label{sec:conclusion}

LongRCA Bench provides 1,140 human-annotated long failure trajectories for
independent role attribution and earliest decisive root-step localization.
RCTA combines original candidate records with explicit handoff evidence for
instruction--execution comparison. On the full benchmark with DS-V4-Flash,
it reaches 24.1\% exact root-step accuracy, compared with 13.2\% for the
strongest evaluated baseline. Despite this improvement, fewer than one
quarter of reference roots are recovered exactly, leaving precise
localization a substantial challenge.

\subsection*{AI Use Statement}

We used GPT-5.6 Sol and Codex for literature discovery, drafting, editing,
and LaTeX formatting, and DeepSeek V4 Flash Preview for annotation-navigation
summaries. We reviewed the AI-assisted content and references and take
responsibility for the final work.

\subsection*{Ethics Statement}

Annotators were informed about the study and participated as part of the
research group's work. We screened the released data for sensitive
information and reviewed the source licenses.

\subsection*{Reproducibility Statement}

Sections~\ref{sec:benchmark}--\ref{sec:experiments} describe benchmark
construction, RCTA, and evaluation. Appendices~\ref{app:mini}
and~\ref{app:annotation} document the sampling manifest and annotation
protocol. Prompt and inference details are provided in
Appendix~\ref{app:prompts}, and scoring rules in
Appendix~\ref{app:scoring_protocols}.

\endgroup
\let\FloatBarrier\maintextSavedFloatBarrier
\FloatBarrier

\bibliography{references}
\bibliographystyle{iclr2027_conference}

\appendix

\section{LongRCA-Mini sampling and reproducibility}
\label{app:mini}

\paragraph{Sampling protocol.}
The sample-size calibration uses master seed 20260820 and 10,000 draws for each
candidate size under equal-quota simple-random sampling. The released 200-QID manifest is generated with the
independent master seed 20260821. It records a SHA-256 hash of the source
dataset and a hash for every selected record. Final manifest selection is
without replacement and uses only source membership, QIDs, and the fixed seed;
no prediction outcomes or reference labels influence which QIDs are selected.
The manifest is not rerolled to improve an observed result and contains 40 QIDs
per source. Full-benchmark predictions from six evaluated methods under DS-V4-Flash
are used to calibrate sample size, not to select individual trajectories.
All methods share the sampled QIDs within each calibration draw and the same
released manifest for the fixed-subset check. Agent Acc follows
the independent explicit-output scoring protocol in Section~\ref{sec:experiments}.

The candidate sizes are 75, 100, 125, 150, 175, 200, 225, 250, 300, 400,
and 600. Equal allocation caps each source at its population size and
redistributes unused slots evenly, with deterministic rounding to preserve the
requested total. All sampled scores retain equal source weights in Macro-5
even when source-size caps yield unequal quotas.

\paragraph{Score errors and calibration criteria.}
Calibration and fixed-manifest fidelity use the same full-benchmark predictions.
For method $m$ and
metric $k$, absolute score error in percentage points (pp) is
\[
D_{m,k}=100\left|S^{\mathrm{Mini}}_{m,k}
-S^{\mathrm{Full}}_{m,k}\right|.
\]
Both scores are fractions aggregated with Macro-5. Full reaggregates all
1,140 predictions per method using equal source weights, which differ
from the original-proportion weighting in Table~\ref{tab:main_results}.
Across the 18 method--metric pairs formed by the six methods and Agent Acc,
Step Acc, and Step $\pm5$, a candidate size passes when
\[
\max_{m,k}\operatorname{Median}(D_{m,k})\leq 2\ \mathrm{pp},
\qquad
\max_{m,k}Q_{0.95}(D_{m,k})\leq 6\ \mathrm{pp},
\]
where each median and quantile is computed over repeated draws for one
method--metric pair before taking the maximum across pairs. These predefined
criteria depend only on absolute score errors.

\begin{figure}[!htbp]
\centering
\includegraphics[width=0.52\linewidth]{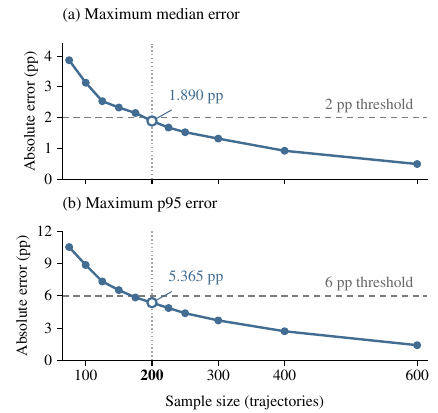}
\caption{Mini calibration with cached predictions and 10,000 source-stratified
draws per size: maximum per-pair median and 95th-percentile absolute Macro-5
errors across 18 method--metric pairs.}
\label{fig:mini_stability}
\end{figure}

\begin{table}[!htbp]
\centering
\caption{Six-method sample-size calibration with equal-quota simple-random
sampling. Quotas follow the source
order SWE-bench Pro, Terminal-Bench 2, TravelPlanner, VitaBench, and WebArena
Verified. For each of 18 method--metric pairs, absolute score errors are
summarized over 10,000 draws. ``Max. median'' and ``Max. p95'' report the
maximum per-pair median and 95th-percentile errors (pp).
``Pass'' requires only max. median $\leq2$ pp and max. p95 $\leq6$ pp.}
\label{tab:mini_sample_sizes}
\normalsize
\fontencoding{OT1}\selectfont
\setlength{\tabcolsep}{5pt}
\begin{tabular}{@{}rlrrc@{}}
\toprule
\textbf{Size} & \textbf{Quotas} &
\textbf{Max. median} & \textbf{Max. p95} & \textbf{Pass} \\
\midrule
75 & 15/15/15/15/15 & 3.865 & 10.532 & No \\
100 & 20/20/20/20/20 & 3.135 & 8.865 & No \\
125 & 25/25/25/25/25 & 2.535 & 7.335 & No \\
150 & 30/30/30/30/30 & 2.327 & 6.532 & No \\
175 & 35/35/35/35/35 & 2.151 & 5.849 & No \\
\textbf{200} & \textbf{40/40/40/40/40} & \textbf{1.890} & \textbf{5.365} & \textbf{Yes} \\
225 & 46/42/46/46/45 & 1.676 & 4.865 & Yes \\
250 & 52/42/52/52/52 & 1.526 & 4.384 & Yes \\
300 & 65/42/65/64/64 & 1.317 & 3.716 & Yes \\
400 & 90/42/90/89/89 & 0.922 & 2.708 & Yes \\
600 & 128/42/161/108/161 & 0.491 & 1.415 & Yes \\
\bottomrule
\end{tabular}
\end{table}

As shown in Figure~\ref{fig:mini_stability} and
Table~\ref{tab:mini_sample_sizes}, $n=200$ is the smallest
tested size satisfying both error criteria; $n=175$ exceeds the median-error
threshold. The empirical p95 uses the nearest-rank definition;
the maximum of per-pair p95 errors is not a simultaneous 18-pair coverage bound.

Equal allocation follows the equal-source Macro-5 objective; calibration
determines the smallest passing size among the tested values for that protocol,
not an optimum over all possible sampling designs or integer sizes.
A supplementary repeat with seed 20260907 and 10,000 draws per candidate size
also yields $n=175$ failing and $n=200$ as the smallest passing tested size.

\paragraph{Fixed-manifest fidelity.}
The fixed-manifest check compares predictions on the released 200 QIDs
with Full under Macro-5. Selection follows the prediction-independent rule
in Section~\ref{sec:mini_construction}.

\begin{table}[!t]
\centering
\caption{Fidelity of the fixed LongRCA-Mini manifest. Each cell reports
\textbf{Full $\rightarrow$ Mini ($|\Delta|$)}: Macro-5 accuracy (\%) and absolute
difference (pp). Across the 18 method--metric comparisons, the mean absolute
deviation is 1.665 pp and the maximum deviation is 3.771 pp. All results use
DS-V4-Flash.}
\label{tab:mini_fidelity}
\normalsize
\fontencoding{OT1}\selectfont
\setlength{\tabcolsep}{7pt}
\renewcommand{\arraystretch}{1.15}
\begin{tabular}{@{}lccc@{}}
\toprule
\textbf{Method} & \textbf{Agent Acc} & \textbf{Step Acc} & \textbf{Step $\pm5$} \\
\midrule
All-at-once & $48.2\rightarrow48.5\ (0.3)$ & $9.9\rightarrow8.5\ (1.4)$ & $23.7\rightarrow21.5\ (2.2)$ \\
Step-by-step & $41.8\rightarrow41.5\ (0.3)$ & $6.5\rightarrow6.5\ (0.0)$ & $19.8\rightarrow17.5\ (2.3)$ \\
Binary search & $40.8\rightarrow39.5\ (1.3)$ & $4.9\rightarrow7.5\ (2.6)$ & $18.1\rightarrow20.0\ (1.9)$ \\
ECHO & $49.4\rightarrow46.5\ (2.9)$ & $17.3\rightarrow13.5\ (3.8)$ & $31.2\rightarrow29.0\ (2.2)$ \\
FALAT & $33.1\rightarrow34.5\ (1.4)$ & $2.7\rightarrow3.0\ (0.3)$ & $15.5\rightarrow18.0\ (2.5)$ \\
\textbf{RCTA} & $\mathbf{60.9}\rightarrow\mathbf{61.0}\ (0.1)$ & $\mathbf{28.3}\rightarrow\mathbf{27.0}\ (1.3)$ & $\mathbf{44.1}\rightarrow\mathbf{41.0}\ (3.1)$ \\
\bottomrule
\end{tabular}
\end{table}

Table~\ref{tab:mini_fidelity} reports a mean absolute deviation of 1.665 pp
and a maximum of 3.771 pp across the 18 method--metric pairs. These deviations
are computed before rounding the displayed scores and apply to this fixed
manifest; calibration quantiles instead summarize repeated samples. The
comparison measures subset-selection discrepancy under shared predictions
and reference labels, not run-to-run inference variance. It supports
preliminary comparisons, ablations, and configuration screening under the
evaluated protocol, without establishing fidelity for future methods.
Close comparisons, per-source claims, headline results, leaderboard
submissions, changed data or scoring protocols, and materially different
method behavior require confirmation on all 1,140 trajectories.

\begin{table}[!htbp]
\centering
\caption{Structural comparison of LongRCA Bench and the fixed
LongRCA-Mini manifest. The KS statistic is the maximum across within-source
comparisons of trajectory length, root-to-end distance, and normalized root
position. History characters are used as a model-independent input-cost proxy.}
\label{tab:mini_structure}
\normalsize
\setlength{\tabcolsep}{4pt}
\begin{tabular}{@{}lrrr@{}}
\toprule
\textbf{Structural quantity} & \textbf{Full} & \textbf{Mini} &
\textbf{Comparison} \\
\midrule
Trajectories & 1,140 & 200 & 17.54\% retained \\
Recorded steps & 178,137 & 32,317 & 18.14\% retained \\
History characters & 207,731,218 & 41,784,984 & 20.11\% retained \\
Macro-5 mean trajectory length & 167.7 & 161.6 & $-6.1$ steps \\
Macro-5 mean root-to-end distance & 104.9 & 104.4 & $-0.5$ steps \\
Macro-5 mean normalized root position & 0.420 & 0.399 & $-0.021$ \\
Maximum within-source KS distance & -- & -- & 0.177 \\
\bottomrule
\end{tabular}
\end{table}

As shown in Table~\ref{tab:mini_structure}, the fixed Mini retains
20.11\% of history characters and 18.14\% of recorded steps. The Macro-5
mean trajectory length differs from Full by 6.1 steps, and the mean
root-to-end distance differs by 0.5 steps. Score fidelity is summarized
in Table~\ref{tab:mini_fidelity}.

\begin{table}[!htbp]
\centering
\caption{Within-source structural audit of the fixed manifest. Entries are
Full/Mini means; input length is measured in thousands of history characters,
and normalized root position is $r^*/(T-1)$. KS is the largest empirical CDF
distance across the four reported characteristics. SWE, TB2, TP, Vita, and WA
denote the five sources in the order used in Table~\ref{tab:mini_sample_sizes}.}
\label{tab:mini_source_structure}
\normalsize
\setlength{\tabcolsep}{3.5pt}
\begin{tabular}{@{}lrrrrr@{}}
\toprule
\textbf{Source} & \textbf{Steps} & \textbf{Input kchars} &
\textbf{Root-to-end} & \textbf{Root position} & \textbf{KS} \\
\midrule
SWE & 287.0/263.0 & 450.2/432.9 & 215.4/218.5 & 0.250/0.181 & 0.177 \\
TB2 & 272.1/265.6 & 297.2/294.9 & 186.9/179.5 & 0.263/0.268 & 0.040 \\
TP & 165.9/167.8 & 163.3/154.1 & 65.9/67.8 & 0.597/0.569 & 0.134 \\
Vita & 54.6/50.8 & 23.2/26.9 & 29.7/26.0 & 0.445/0.473 & 0.134 \\
WA & 58.8/60.8 & 131.2/135.8 & 26.8/30.1 & 0.543/0.502 & 0.147 \\
\bottomrule
\end{tabular}
\end{table}

As shown in Table~\ref{tab:mini_source_structure}, the maximum
within-source KS distance ranges from 0.040 for Terminal Bench 2 to
0.177 for SWE-bench Pro. The largest change in mean trajectory length
occurs in SWE-bench Pro, from 287.0 to 263.0 steps.

\begin{table}[!htbp]
\centering
\caption{Per-source Full/Mini accuracy (\%) for the fixed manifest. Each
cell reports the full source score followed by the Mini source score. Source
abbreviations follow Table~\ref{tab:mini_source_structure}. These values are an
audit of source-level variation, not evidence that 40 trajectories support
definitive per-source comparisons.}
\label{tab:mini_per_source}
\normalsize
\fontencoding{OT1}\selectfont
\setlength{\tabcolsep}{3.5pt}
\renewcommand{\arraystretch}{1.06}
\begin{tabular}{@{}llrrrrr@{}}
\toprule
\textbf{Method} & \textbf{Metric} & \textbf{SWE} & \textbf{TB2} &
\textbf{TP} & \textbf{Vita} & \textbf{WA} \\
\midrule
All-at-once & Agent Acc & 47.7/42.5 & 61.9/62.5 & 7.2/7.5 & 82.4/85.0 & 41.8/45.0 \\
            & Step Acc & 1.6/2.5 & 7.1/5.0 & 4.5/5.0 & 19.4/20.0 & 16.9/10.0 \\
            & Step $\pm5$ & 4.7/7.5 & 21.4/20.0 & 15.2/10.0 & 41.7/42.5 & 35.6/27.5 \\
\midrule
Step-by-step & Agent Acc & 38.3/30.0 & 40.5/40.0 & 2.8/5.0 & 83.3/85.0 & 44.1/47.5 \\
             & Step Acc & 0.8/2.5 & 4.8/5.0 & 3.5/0.0 & 12.0/12.5 & 11.3/12.5 \\
             & Step $\pm5$ & 10.2/7.5 & 16.7/15.0 & 13.0/10.0 & 30.6/27.5 & 28.8/27.5 \\
\midrule
Binary search & Agent Acc & 39.1/27.5 & 45.2/45.0 & 7.4/7.5 & 82.4/85.0 & 29.9/32.5 \\
              & Step Acc & 0.8/2.5 & 4.8/2.5 & 2.0/7.5 & 12.0/17.5 & 5.1/7.5 \\
              & Step $\pm5$ & 3.9/5.0 & 23.8/22.5 & 9.5/15.0 & 32.4/35.0 & 20.9/22.5 \\
\midrule
ECHO & Agent Acc & 45.3/32.5 & 59.5/60.0 & 7.2/5.0 & 82.4/85.0 & 52.5/50.0 \\
     & Step Acc & 7.8/7.5 & 21.4/22.5 & 8.9/7.5 & 22.2/20.0 & 26.0/10.0 \\
     & Step $\pm5$ & 18.0/17.5 & 38.1/37.5 & 18.8/17.5 & 42.6/45.0 & 38.4/27.5 \\
\midrule
FALAT & Agent Acc & 32.0/37.5 & 28.6/30.0 & 6.6/10.0 & 79.6/80.0 & 18.6/15.0 \\
      & Step Acc & 2.3/5.0 & 2.4/2.5 & 2.9/2.5 & 3.7/2.5 & 2.3/2.5 \\
      & Step $\pm5$ & 10.9/10.0 & 16.7/17.5 & 10.1/10.0 & 26.9/32.5 & 13.0/20.0 \\
\midrule
RCTA & Agent Acc & 63.3/52.5 & 57.1/55.0 & 41.2/47.5 & 82.4/85.0 & 60.5/65.0 \\
     & Step Acc & 38.3/32.5 & 26.2/25.0 & 18.8/22.5 & 25.0/32.5 & 33.3/22.5 \\
     & Step $\pm5$ & 46.9/40.0 & 45.2/42.5 & 29.9/27.5 & 47.2/52.5 & 51.4/42.5 \\
\bottomrule
\end{tabular}
\end{table}

As shown in Table~\ref{tab:mini_per_source}, RCTA Step Acc changes
from 38.3\% to 32.5\% on SWE-bench Pro and from 25.0\% to 32.5\% on
VitaBench. These source-level shifts describe variation within the fixed
manifest; each Mini source contains only 40 trajectories.

For direct comparison with the full benchmark's natural source mixture, the
Mini source scores can instead be post-stratified as
$S_{\mathrm{original}}=\sum_{b=1}^{5}(N_b/1140)S_b$.

\begin{table}[!htbp]
\centering
\caption{Original-mixture fidelity audit. Cells report Full $\rightarrow$
post-stratified Mini ($|\Delta|$), in accuracy percentages and pp.
Both sides use the original source weights, not Macro-5.}
\label{tab:mini_weighted}
\normalsize
\fontencoding{OT1}\selectfont
\setlength{\tabcolsep}{6pt}
\begin{tabular}{@{}lccc@{}}
\toprule
\textbf{Method} & \textbf{Agent Acc} & \textbf{Step Acc} & \textbf{Step $\pm5$} \\
\midrule
All-at-once & $26.2\rightarrow26.6\ (0.4)$ & $7.6\rightarrow6.9\ (0.7)$ & $19.9\rightarrow15.9\ (4.0)$ \\
Step-by-step & $22.2\rightarrow23.3\ (1.1)$ & $5.3\rightarrow3.6\ (1.7)$ & $16.9\rightarrow14.3\ (2.7)$ \\
Binary search & $23.0\rightarrow22.4\ (0.6)$ & $3.4\rightarrow7.7\ (4.3)$ & $13.3\rightarrow17.2\ (3.9)$ \\
ECHO & $27.5\rightarrow24.7\ (2.9)$ & $13.2\rightarrow9.6\ (3.5)$ & $24.7\rightarrow22.4\ (2.3)$ \\
FALAT & $19.0\rightarrow21.2\ (2.2)$ & $2.8\rightarrow2.8\ (0.0)$ & $12.5\rightarrow14.0\ (1.5)$ \\
RCTA & $51.1\rightarrow54.6\ (3.5)$ & $24.1\rightarrow24.7\ (0.5)$ & $37.4\rightarrow34.2\ (3.2)$ \\
\bottomrule
\end{tabular}
\end{table}

In Table~\ref{tab:mini_weighted}, Full and Mini use the original
source proportions as their aggregation weights. This secondary estimand is separate from
Macro-5 in Table~\ref{tab:mini_fidelity}; the two types of aggregate should
not be compared as if they used the same source weights.

\begingroup
\rightskip=0pt plus 2em
\paragraph{Reproduction artifacts.}
The release includes {\urlstyle{same}\nolinkurl{scripts/sample_equal_benchmark_dataset.py}},
{\urlstyle{same}\nolinkurl{scripts/analyze_longrca_six_method_calibration.py}},
{\urlstyle{same}\nolinkurl{scripts/analyze_longrca_mini_fidelity.py}}, and
{\urlstyle{same}\nolinkurl{scripts/plot_longrca_six_method_calibration.py}}, together with the fixed QID
manifest, repeated-draw indices and scores, and per-size and per-method--metric
results. Input hashes and the Python,
NumPy, and Matplotlib versions are recorded in the analysis manifests.
\par
\endgroup

The calibration code verifies identical 1,140-QID coverage across the six
methods, exact quotas and no within-draw duplicate QIDs, and identical draw
indices and scores under same-seed replay. The sampling function accepts only
source population sizes, sample size, repetition count, and seed; prediction
scores are joined after drawing. The fixed-manifest fidelity check reuses
the released 200 QIDs without modifying their selection.
\section{Complete cross-backbone results on LongRCA-Mini}
\label{app:mini_backbones}

In the complete results in Table~\ref{tab:mini_backbones_full}, each source contributes 40 trajectories,
so aggregate accuracies equal Macro-5. Joint Exact and Joint $\pm5$ require
both correct role attribution and the corresponding step criterion; Step MAE
summarizes valid-output absolute root-step error. The entries in Table~\ref{tab:mini_backbones_full} are descriptive point estimates rather than repeated-run averages; paired uncertainty analyses and significance tests are reported separately below.

\begin{table}[!htbp]
\centering
\caption{Complete six-metric record for the fixed 200-trajectory
cross-backbone experiment. Accuracies are Macro-5 percentages; Step MAE is
the reported valid-output mean absolute root-step error in steps.
Bold marks the best value within each backbone and metric (including ties).
Shading identifies RCTA. Evaluation settings and prediction coverage are
detailed in the accompanying text.}
\label{tab:mini_backbones_full}
\normalsize
\fontencoding{OT1}\selectfont
\setlength{\tabcolsep}{5pt}
\renewcommand{\arraystretch}{1.08}
\begin{tabular}{@{}lrrrrrr@{}}
\toprule
\textbf{Method} & \textbf{Agent Acc} $\uparrow$ & \textbf{Step Acc} $\uparrow$ &
\textbf{Step $\pm5$} $\uparrow$ & \textbf{Joint Exact} $\uparrow$ &
\textbf{Joint $\pm5$} $\uparrow$ & \textbf{MAE} $\downarrow$ \\
\midrule
\multicolumn{7}{l}{\textbf{DS-V4-Flash}} \\
\midrule
All-at-once & 42.5 & 7.5 & 20.5 & 7.5 & 13.0 & 44.915 \\
Step-by-step & 38.5 & 6.5 & 18.0 & 6.5 & 10.0 & 41.340 \\
Binary search & 40.5 & 7.5 & 20.5 & 7.5 & 12.0 & 43.465 \\
ECHO & 45.0 & 13.0 & 28.0 & 13.0 & 18.0 & 40.930 \\
FALAT & 35.5 & 2.5 & 17.5 & 2.5 & 10.5 & 53.685 \\
\rowcolor[gray]{0.96}
\textbf{RCTA} & \textbf{60.5} & \textbf{27.0} & \textbf{41.0} & \textbf{27.0} & \textbf{34.5} & \textbf{28.360} \\
\midrule
\multicolumn{7}{l}{\textbf{DS-V4-Pro}} \\
\midrule
All-at-once & 50.0 & 11.5 & 29.5 & 11.5 & 19.5 & 36.200 \\
Step-by-step & 47.5 & 10.5 & 23.0 & 10.5 & 16.0 & 41.865 \\
Binary search & 45.5 & 10.0 & 28.5 & 10.0 & 18.0 & 34.555 \\
ECHO & 53.5 & 18.0 & 35.5 & 18.0 & 25.5 & 36.540 \\
FALAT & 48.0 & 14.5 & 23.5 & 14.5 & 20.5 & 52.370 \\
\rowcolor[gray]{0.96}
\textbf{RCTA} & \textbf{61.0} & \textbf{28.0} & \textbf{43.0} & \textbf{28.0} & \textbf{37.0} & \textbf{31.365} \\
\midrule
\multicolumn{7}{l}{\textbf{GPT-5.6-sol}} \\
\midrule
All-at-once & 53.0 & 17.0 & 34.0 & 17.0 & 26.0 & 42.690 \\
Step-by-step & 49.0 & 13.0 & 26.0 & 13.0 & 21.0 & 40.635 \\
Binary search & 51.5 & 15.5 & 31.5 & 15.5 & 25.5 & 32.710 \\
ECHO & 55.5 & 18.5 & 32.0 & 18.5 & 24.5 & 44.305 \\
FALAT & 52.0 & 14.0 & 29.0 & 14.0 & 21.5 & 39.820 \\
\rowcolor[gray]{0.96}
\textbf{RCTA} & \textbf{64.5} & \textbf{30.0} & \textbf{42.0} & \textbf{29.5} & \textbf{38.0} & \textbf{31.685} \\
\midrule
\multicolumn{7}{l}{\textbf{Kimi K3}} \\
\midrule
All-at-once & 54.5 & 11.0 & 29.5 & 10.5 & 20.5 & 34.225 \\
Step-by-step & 44.0 & 9.5 & 23.5 & 9.5 & 16.0 & 48.855 \\
Binary search & 42.0 & 12.0 & 28.0 & 12.0 & 20.5 & 40.145 \\
ECHO & 55.0 & 20.5 & 37.0 & 20.5 & 28.0 & \textbf{30.760} \\
FALAT & 52.0 & 18.0 & 30.0 & 18.0 & 24.0 & 41.265 \\
\rowcolor[gray]{0.96}
\textbf{RCTA} & \textbf{61.5} & \textbf{31.5} & \textbf{41.5} & \textbf{31.0} & \textbf{37.5} & 31.995 \\
\midrule
\multicolumn{7}{l}{\textbf{Qwen3.5-35B}} \\
\midrule
All-at-once & \textbf{51.0} & 3.5 & 18.0 & 2.5 & 11.0 & \textbf{42.550} \\
Step-by-step & 46.5 & 4.5 & 18.0 & 3.0 & 12.0 & 44.865 \\
Binary search & 40.0 & 5.0 & 18.0 & 4.0 & 12.5 & 43.560 \\
ECHO & 48.5 & 10.0 & \textbf{26.5} & 9.0 & 19.5 & 44.650 \\
FALAT & 41.0 & 11.0 & 20.0 & 11.0 & 15.5 & 51.345 \\
\rowcolor[gray]{0.96}
\textbf{RCTA} & 48.5 & \textbf{13.5} & 25.0 & \textbf{13.5} & \textbf{21.0} & 45.330 \\
\bottomrule
\end{tabular}
\end{table}

\paragraph{Evaluation protocol and prediction coverage.}
DS-V4-Flash, DS-V4-Pro, GPT-5.6 Sol, and Kimi K3 each have
200 unique predictions per method on the shared 40-per-source sample,
providing 4,800 method--trajectory records for the paired analysis below.
Kimi K3 is supported by complete per-trajectory predictions,
with the same fixed accuracy denominators as the other backbones.
The table retains the MAE exception on Kimi K3: ECHO obtains 30.760 steps,
compared with 31.995 for RCTA. The paired RCTA--ECHO MAE difference is
1.235 steps (95\% CI: $[-6.975,9.235]$), so the point-estimate reversal
is not a resolved MAE difference. Exact-localization gains need not imply
superiority on every reported measure.

Qwen3.5-35B-A3B has 1,200 valid method--trajectory records (200 per method,
40 per source), with no missing, duplicate, or parse-failed predictions.
Its scores use explicit role and root-step outputs;
RCTA reads \texttt{root\_cause.agent/step}.
On this backbone, RCTA also leads Joint Exact (13.5\%) and Joint $\pm5$
(21.0\%). Although ECHO has higher Step $\pm5$ (26.5\% versus 25.0\%),
RCTA has higher Joint $\pm5$ (21.0\% versus 19.5\%), indicating more cases
where role and tolerant step predictions are simultaneously correct.
This aggregate observation alone does not establish a mechanism of improved
role--step consistency.

\paragraph{Paired Step Acc comparisons.}
All paired comparisons retain the same 200 trajectories, with invalid
predictions counted as errors. Table~\ref{tab:mini_paired_step} reports
pointwise 95\% source-stratified paired-bootstrap intervals from 20,000
resamples and two-sided exact McNemar tests. Holm-20 adjusts the 20 Step Acc
comparisons of RCTA with five baselines on four backbones; Holm-40 adjusts
the combined 40-comparison family of Agent Acc and Step Acc. Both are
reported to show sensitivity to the endpoint family, rather than treating
pointwise intervals as simultaneous guarantees. These analyses use one
retained prediction per method and backbone; resampling trajectories does
not capture inference-time variability. Qwen3.5-35B-A3B is not included in this
paired analysis.

\begin{table}[!htbp]
\centering
\caption{Paired RCTA--ECHO Step Acc differences on LongRCA-Mini ($n=200$).
$\Delta$ and pointwise 95\% intervals are in percentage points; W/L counts
RCTA-only/ECHO-only correct predictions. Holm-20 adjusts the Step Acc
family; Holm-40 jointly adjusts Agent Acc and Step Acc.}
\label{tab:mini_paired_step}
\small
\setlength{\tabcolsep}{4pt}
\begin{tabular}{@{}lrrrrr@{}}
\toprule
Backbone & $\Delta$ & 95\% CI & W/L & Holm-20 $p$ & Holm-40 $p$ \\
\midrule
DS-V4-Flash & $+14.0$ & $[7.5, 20.5]$ & 39/11 & 0.00072 & 0.00217 \\
DS-V4-Pro & $+10.0$ & $[3.5, 17.0]$ & 36/16 & 0.00779 & 0.06007 \\
GPT-5.6 Sol & $+11.5$ & $[4.5, 19.0]$ & 40/17 & 0.00640 & 0.04162 \\
Kimi K3 & $+11.0$ & $[5.0, 17.0]$ & 31/9 & 0.00322 & 0.01155 \\
\bottomrule
\end{tabular}
\end{table}

For DS-V4-Pro, the adjusted Step Acc comparison changes from $p=.00779$
under Holm-20 to $p=.060$ under Holm-40; the other three RCTA--ECHO
comparisons remain below $.05$ under both families. A source-stratified
task-cluster bootstrap that keeps repeated attempts together also gives
positive pointwise RCTA--ECHO Step Acc intervals on all four backbones
(184 task groups among 200 trajectories). These checks concern the evaluated
samples and settings, not unseen benchmarks or backbones.

\paragraph{Sensitivity to source composition.}
For Table~\ref{tab:mini_leave_benchmark_out}, we average the four model
outcomes within each trajectory and then average over the retained
benchmarks. Excluding each benchmark in turn leaves a mean gain of
8.3--13.0 percentage points on the remaining Mini trajectories. This is a
sensitivity of the observed aggregate to source composition. Bootstrap
resampling keeps a trajectory's model outcomes together and preserves the
retained source proportions; the five intervals are pointwise, not
simultaneous coverage guarantees.

\begin{table}[!htbp]
\centering
\caption{Leave-one-benchmark-out sensitivity of the four-backbone mean
RCTA--ECHO Step Acc gain. Each row retains 160 trajectories from four
benchmarks; differences and intervals are in percentage points.}
\label{tab:mini_leave_benchmark_out}
\small
\setlength{\tabcolsep}{8pt}
\begin{tabular}{@{}lrr@{}}
\toprule
Excluded benchmark & Mean gain & 95\% CI \\
\midrule
SWE-bench Pro & $+8.3$ & $[3.6, 13.1]$ \\
Terminal-Bench 2 & $+13.0$ & $[8.6, 17.5]$ \\
TravelPlanner & $+11.4$ & $[6.4, 16.4]$ \\
VitaBench & $+13.0$ & $[7.8, 18.1]$ \\
WebArena Verified & $+12.5$ & $[7.3, 17.8]$ \\
\bottomrule
\end{tabular}
\end{table}

\paragraph{Complementary correct localizations.}
\begin{figure}[!htbp]
\centering
\includegraphics[width=0.90\linewidth]{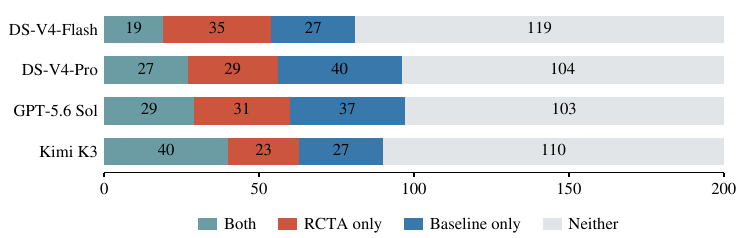}
\caption{Exact-root correctness on LongRCA-Mini ($n=200$ per backbone).
Both: RCTA and at least one of five baselines are correct; RCTA only:
RCTA is correct and all five baselines are wrong; Baseline only: RCTA is
wrong and at least one baseline is correct; Neither: all six methods are
wrong. Segment labels are trajectory counts.}
\label{fig:method_success_overlap}
\end{figure}

Figure~\ref{fig:method_success_overlap} compares RCTA with the union of
five baseline correct sets separately for each backbone on LongRCA-Mini.
Both directions of complementarity are retained: RCTA correctly localizes
23--35 trajectories per backbone that all five baselines miss, while the
baselines correctly localize 27--40 that RCTA misses. RCTA's total correct
count combines the ``Both'' and ``RCTA only'' segments; the latter measures
unique coverage relative to the baselines, not total accuracy.
The union is a gold-informed set summary, not the performance of a deployed
ensemble or a cost-matched competitor. The four rows reuse the same
200 Mini trajectories and must not be added to count unique instances.

Across all four backbones, 26 trajectories have at least one correct RCTA
prediction and no correct prediction from any of the 20 baseline
method--backbone pairs. Conversely, 43 have a correct baseline prediction
and no correct RCTA prediction; 75 have successes from both method sets,
and 56 have none. These counts summarize differently sized sets of
observed predictions, not an ensemble comparison.
Among the 26 RCTA-only trajectories, 18 have non-handoff reference roots.
Inspecting all recorded role names further identifies 15 trajectories
without any explicit handoff marker: eight from TravelPlanner, four from
VitaBench, and three from WebArena Verified. The absence of the explicit
marker does not establish the absence of implicit coordination or identify
which component produced the successful attribution.

\paragraph{Illustrative roots with and without explicit handoffs.}
Two selected 162-step trajectories illustrate the observed coverage.
In \texttt{swe\_bench\_pro\_\_071}, the reference root is a handoff at step 27:
the instruction requests a new \texttt{repoquery} parser but omits connecting
it to \texttt{parseInstalledPackages} for the six-field output.
In \texttt{travelplanner\_\_458}, the reference root is step 84, where the
accommodation agent proposes using estimated industry-average prices after
its tool results fail to provide suitable listings for the required cities.
This trajectory contains no explicit handoff marker. RCTA exactly matches
the reference step on both cases with all four backbones, whereas none of
the 20 baseline method--backbone results does. The cases illustrate distinct
root structures; they do not estimate the prevalence of task-specific
failure types or constitute a controlled comparison of components.

\section{Supplementary comparison with Who\&When}
\label{app:whowhen}
\label{sec:whowhen_comparison}

We evaluate all six methods on the complete Who\&When dataset
\citep{whowhen} using DS-V4-Flash. The dataset contains 184 trajectories:
126 from algorithm-generated systems and 58 from hand-crafted systems. We
compare these results with the full LongRCA evaluation, weighting each
trajectory equally within its dataset.

The five baselines, as shown in Figure~\ref{fig:whowhen_comparison}, obtain
30.4--36.4\% Step Acc on Who\&When and 2.8--13.2\% on LongRCA. RCTA obtains
41.8\% and 24.1\%, respectively. It has the highest observed accuracy on all
three metrics on LongRCA. On Who\&When, it leads in Agent Acc and Step Acc,
while All-at-once leads in Step $\pm5$.

All six methods have lower accuracy on LongRCA for each of the three metrics
in this evaluation. The median trajectory length is 10 steps in Who\&When
and 145 in LongRCA, so a five-step localization error has a different relative
scale in the two datasets. We use Step Acc as the primary localization
comparison and report Step $\pm5$ as a complementary measure. The datasets
also differ in domains and workflows, so the cross-dataset accuracy gaps do
not isolate the effect of trajectory length.

\section{Released data schema}
\label{app:schema}

Normalization maps each retained execution to an ordered sequence of records
covering multi-agent dialogue, planning, tool calls, observations, browser or
terminal interactions, and evaluator feedback. Records preserve original
content and roles, stable 0-based indices, and available tool and verifier
metadata. Temporal order is preserved without synthesizing intermediate
actions or subdividing source events into artificial steps. Annotators,
reference labels, and evaluated methods share these indices within each
trajectory; event granularity can still differ across sources.

Source evaluator outcomes guide failure selection and human annotation.
Evaluated methods receive only \texttt{question\_ID} and \texttt{history},
which includes the task instruction. Reference annotations are withheld
during prediction, and only the responsible-role and root-step fields are
scored. This field boundary does not remove evaluator or verifier feedback
already recorded within the history.

\begin{table}[!htbp]
\centering
\caption{Released LongRCA Bench data schema. Evaluated methods receive
\texttt{question\_ID} and \texttt{history}, which includes the task instruction.
The three reference annotation fields are withheld during prediction.}
\label{tab:schema}
\normalsize
\setlength{\tabcolsep}{4pt}
\begin{tabularx}{\linewidth}{@{}>{\raggedright\arraybackslash}p{0.26\linewidth}>{\raggedright\arraybackslash}X@{}}
\toprule
\textbf{Field} & \textbf{Meaning} \\
\midrule
\texttt{question\_ID} & Source-prefixed trajectory identifier \\
\texttt{history} & Ordered trajectory records containing \texttt{step},
\texttt{name}, \texttt{role}, and \texttt{content}; includes the task instruction \\
\texttt{mistake\_agent} & Reference responsible role, scored independently of
the root-cause step \\
\texttt{mistake\_step} & Reference 0-based root-cause step, scored independently
of the responsible role \\
\texttt{mistake\_reason} & Trajectory-grounded annotation rationale, withheld
during prediction and not scored \\
\bottomrule
\end{tabularx}

\end{table}

\section{Source execution provenance}
\label{app:source_provenance}

Table~\ref{tab:source_runs} summarizes task domains and agent organizations;
Table~\ref{tab:source_provenance} records workflow implementations and
trajectory-generator assignments. Fixed-role teams, group-chat coordination,
and sequential workflows differ in their role sets, handoff patterns, and
placement of plans, tool results, and verification messages.

\begin{table}[!htbp]
\centering
\caption{Task domains and agent organizations of the retained LongRCA Bench
trajectories. Counts are reported by source.}
\label{tab:source_runs}

\normalsize
\setlength{\tabcolsep}{3pt}
\renewcommand{\arraystretch}{1.12}

\begin{tabularx}{\linewidth}{
@{}
l
r
>{\raggedright\arraybackslash}p{1.45in}
>{\raggedright\arraybackslash}X
@{}
}
\toprule
\textbf{Source} &
\textbf{$N$} &
\textbf{Task domain} &
\textbf{Agent organization} \\
\midrule

SWE-bench Pro &
128 &
Software repair &
Diagnosis--execution--verification team \\

Terminal Bench 2 &
42 &
Terminal tasks &
Diagnosis--execution--verification team \\

TravelPlanner &
685 &
Travel planning &
Magentic-One group chat \\

VitaBench &
108 &
Service-oriented tool use &
Action expert--computer terminal workflow \\

WebArena Verified &
177 &
Web interaction &
Planner--web-surfer--critic workflow \\

\midrule
\textbf{Total} &
\textbf{1,140} &
\multicolumn{2}{l}{\textbf{5 task domains}} \\

\bottomrule
\end{tabularx}
\end{table}

\begin{table}[!htbp]
\centering
\caption{Execution provenance and generator counts for LongRCA Bench. The
SWE-bench Pro and Terminal-Bench 2 execution records specify their custom role workflow
and generator, but no separate third-party scaffold package. Generator columns
correspond to MiniMax-M2.5 (M2.5), Kimi-K2.5 (K2.5), and Qwen3.5-Plus (Q3.5+).
A dash indicates that the source contains no trajectory from that generator.}
\label{tab:source_provenance}
\normalsize
\setlength{\tabcolsep}{3.5pt}
\begin{tabularx}{\linewidth}{@{}l r >{\raggedright\arraybackslash}X r r r@{}}
\toprule
\textbf{Source} & \textbf{$N$} & \textbf{Workflow} &
\textbf{M2.5} & \textbf{K2.5} & \textbf{Q3.5+} \\
\midrule
SWE-bench Pro & 128 & Custom \texttt{DiagnostAgent}--\texttt{ActionAgent}--\texttt{JudgeAgent} team & 128 & -- & -- \\
Terminal-Bench 2 & 42 & Custom \texttt{DiagnostAgent}--\texttt{ActionAgent}--\texttt{JudgeAgent} team & 42 & -- & -- \\
TravelPlanner & 685 & AutoGen \texttt{MagenticOneGroupChat} with specialist roles & 265 & 260 & 160 \\
VitaBench & 108 & \texttt{Action\_Expert}--\texttt{Computer\_terminal} workflow & 69 & 16 & 23 \\
WebArena Verified & 177 & AutoGen \texttt{Planner}--\texttt{WebSurfer}--\texttt{Critic} controller & 177 & -- & -- \\
\midrule
\textbf{Total} & \textbf{1,140} & & \textbf{681} & \textbf{276} & \textbf{183} \\
\bottomrule
\end{tabularx}

\end{table}

Source representation is uneven: TravelPlanner contributes 685 trajectories
(60.1\%), while the other four sources together contribute 455 (39.9\%).
MiniMax-M2.5 generated 681 trajectories, Kimi-K2.5 generated 276, and
Qwen3.5-Plus generated 183. Only TravelPlanner and VitaBench contain
trajectories from all three generators. These counts describe trajectory
generation rather than the inference backbones used for attribution.

\section{Localization by source and log characteristics}
\label{app:stratified}

\begin{figure}[!htbp]
\centering
\includegraphics[width=\linewidth]{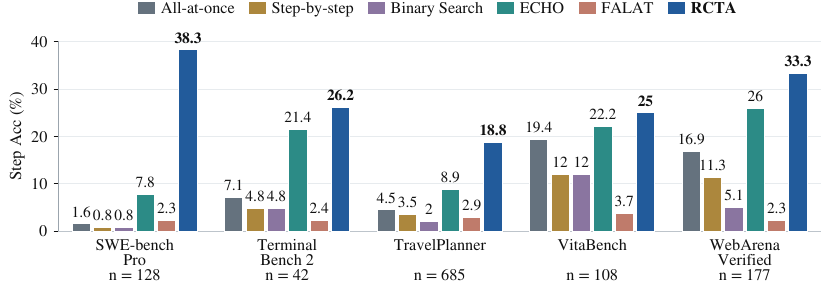}
\caption{Step Acc (\%) across five sources using DS-V4-Flash. All six methods
use the same complete source subsets ($n$ below source labels; total 1,140);
invalid step outputs count as errors.}
\label{fig:benchmark_accuracy}
\end{figure}

\begin{table}[!htbp]
\centering
\caption{Full-benchmark Step Acc (\%) by source, underlying
Figure~\ref{fig:benchmark_accuracy}. Source sample sizes are 128, 42, 685,
108, and 177, respectively. Every method uses the complete corresponding
source subset; bold marks the highest observed accuracy in each source.}
\label{tab:full_source_step_accuracy}
\normalsize
\fontencoding{OT1}\selectfont
\setlength{\tabcolsep}{5pt}
\renewcommand{\arraystretch}{1.12}
\begin{tabular}{@{}lrrrrr@{}}
\toprule
\textbf{Method} & \shortstack{\textbf{SWE-bench}\\\textbf{Pro}} &
\shortstack{\textbf{Terminal}\\\textbf{Bench 2}} & \textbf{TravelPlanner} &
\textbf{VitaBench} & \shortstack{\textbf{WebArena}\\\textbf{Verified}} \\
\midrule
All-at-once & 1.6 & 7.1 & 4.5 & 19.4 & 16.9 \\
Step-by-step & 0.8 & 4.8 & 3.5 & 12.0 & 11.3 \\
Binary search & 0.8 & 4.8 & 2.0 & 12.0 & 5.1 \\
ECHO & 7.8 & 21.4 & 8.9 & 22.2 & 26.0 \\
FALAT & 2.3 & 2.4 & 2.9 & 3.7 & 2.3 \\
\textbf{RCTA} & \textbf{38.3} & \textbf{26.2} & \textbf{18.8} & \textbf{25.0} & \textbf{33.3} \\
\bottomrule
\end{tabular}

\end{table}

\begin{table}[!htbp]
\centering
\caption{Step Acc (\%) across the five shared trajectory-length groups in
Figure~\ref{fig:length_accuracy}. The groups contain 290, 211, 212, 215,
and 212 trajectories, respectively. Bold marks the highest observed accuracy
in each group.}
\label{tab:stratified_length}
\normalsize
\fontencoding{OT1}\selectfont
\setlength{\tabcolsep}{6pt}
\renewcommand{\arraystretch}{1.12}
\begin{tabular}{@{}lrrrrr@{}}
\toprule
\textbf{Method} & \textbf{10--63} & \textbf{64--129} &
\textbf{130--172} & \textbf{173--220} & \textbf{221--728} \\
\midrule
All-at-once & 21.4 & 6.2 & 2.8 & 0.9 & 1.9 \\
Step-by-step & 13.4 & 3.3 & 2.8 & 3.3 & 0.5 \\
Binary search & 7.2 & 3.8 & 0.5 & 2.8 & 1.4 \\
ECHO & 27.9 & 11.4 & 7.1 & 7.9 & 6.1 \\
FALAT & 4.1 & 3.8 & 0.0 & 3.3 & 2.4 \\
\textbf{RCTA} & \textbf{34.8} & \textbf{18.5} & \textbf{21.2} & \textbf{18.6} & \textbf{23.6} \\
\bottomrule
\end{tabular}

\end{table}

As shown in Tables~\ref{tab:full_source_step_accuracy} and~\ref{tab:stratified_length},
RCTA has the highest observed Step Acc in each source and each length group.
These values underlie Figures~\ref{fig:benchmark_accuracy} and~\ref{fig:length_accuracy}. Both use the same full-benchmark
predictions as Table~\ref{tab:main_results}. Each accuracy denominator includes
all trajectories assigned to the corresponding source or length group.

Within the longest group, 170 of 212 trajectories (80.2\%) have 221--400
steps. The RCTA--ECHO Step Acc difference in this group is 17.5 percentage
points; it does not isolate performance on the extreme longest trajectories.

\paragraph{Shared length partition.}
The partition is constructed from one length per unique trajectory, with no
prediction outcomes as inputs. Dynamic programming selects four cut points
between distinct observed lengths to minimize the sum of squared deviations
of group sizes from $1{,}140/5$, while keeping equal lengths together. The
selected boundaries are 63, 129, 172, and 220 steps. A concentration of 145
trajectories at exactly 63 steps limits how closely the groups can approach
equal sizes. Five groups provide a descriptive resolution for the main figure.

\paragraph{Sensitivity to the number of groups.}
Repeating this length-only partition with four and six groups gives group
sizes of 290/281/283/286 and 145/198/199/200/198/200, respectively. RCTA
has the highest observed Step Acc in every group under all three choices,
whereas local curve shapes vary. This sensitivity check supports the
method comparison across these partitions while motivating caution when
interpreting individual fluctuations. Source domains and workflows differ
across the resulting groups.

\subsection{Within-source localization patterns}
\label{app:within_source}

Using the full-benchmark DS-V4-Flash predictions for all six methods, we
examine localization within each source along two dimensions: root-to-end
distance, $D=T-1-r^*$, and log text volume, measured by the total number of
Unicode characters in history content, including whitespace. The former
counts recorded activity after the reference root; the latter measures the
amount of log text.

\begin{figure}[!htbp]
\centering
\includegraphics[width=\linewidth,height=0.80\textheight,keepaspectratio]{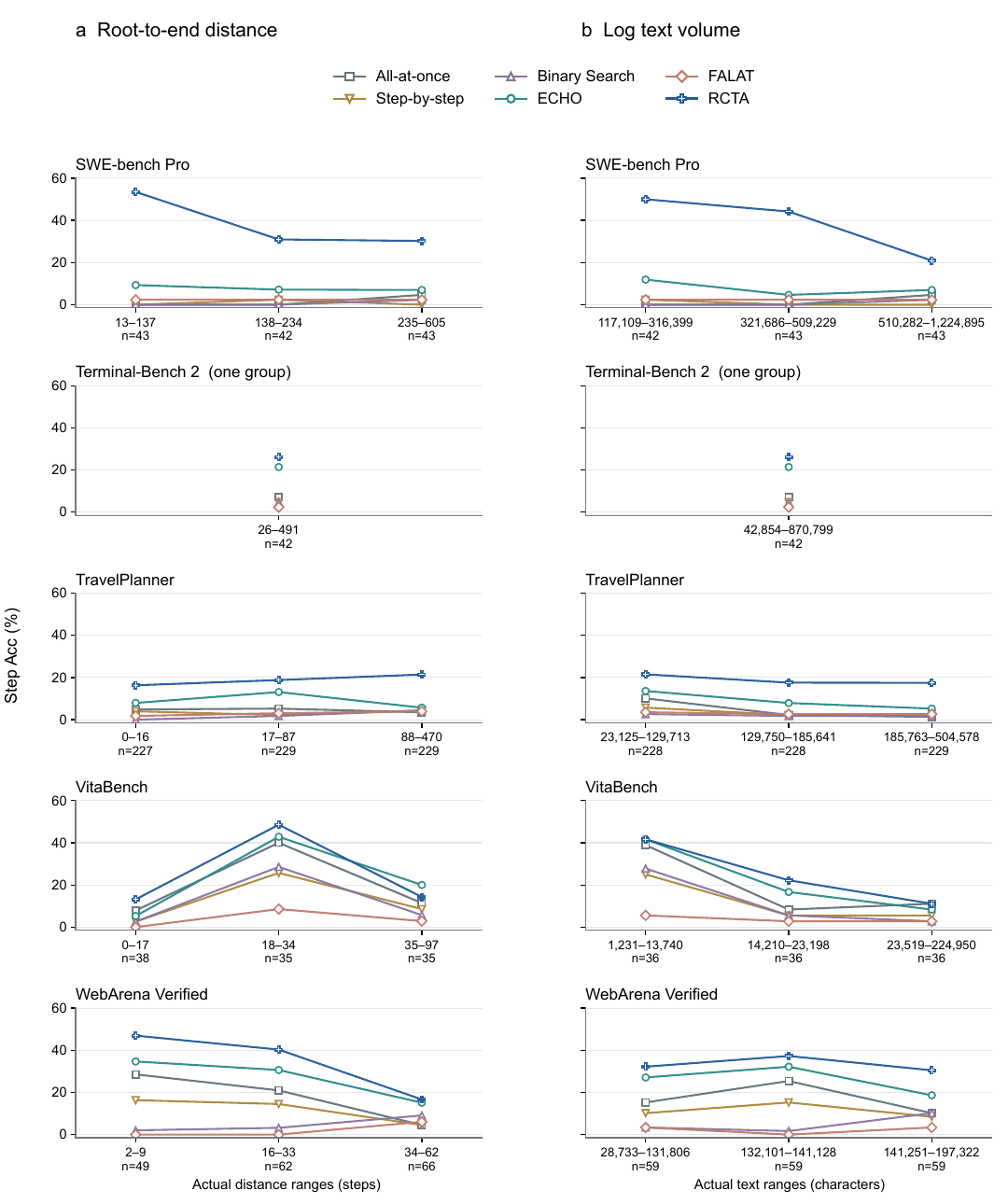}
\caption{Within-source Step Acc with DS-V4-Flash by (a) root-to-end distance
and (b) log text volume. All six methods share the trajectories in each
group. Labels give actual ranges and sample sizes; unequal-width groups
are equally spaced, and connecting lines are visual guides.
Terminal-Bench 2 remains a single group under the minimum-size rule.}
\label{fig:within_source_localization}
\end{figure}

For each source and dimension, we choose the largest feasible number of
groups up to three, with at least 30 trajectories per group and equal
values kept together. Cut points minimize the sum of squared deviations
from equal group sizes using only the grouping variable. All six methods
share each group's trajectories, and invalid predictions count as errors.
Terminal-Bench 2 has 42 trajectories and remains a single group.
Figure~\ref{fig:within_source_localization} shows Step Acc;
Table~\ref{tab:within_source_distance_text} reports both localization metrics.
The accompanying analysis data include Wilson 95\% intervals for group
accuracies and Newcombe 95\% intervals for each method's difference between
the highest and lowest groups \citep{newcombe1998difference}. These intervals are
pointwise and conditional on the fixed groups.

Root-to-end distance shows heterogeneous patterns. On VitaBench, all six
methods attain their highest Step Acc and Step $\pm5$ in the middle group;
on TravelPlanner, differences between the highest and lowest groups have
mixed signs across methods and metrics. Log text volume shows a more consistent contrast in
these two sources: all six methods score lower in the highest-character
group than in the lowest-character group on both metrics. Step Acc changes
range from $-8.8$ to $-0.9$ pp on TravelPlanner and from $-33.3$ to $-2.8$ pp
on VitaBench, with some intermediate-group reversals. Patterns on SWE-bench
Pro and WebArena Verified vary by method. These source-specific associations
do not establish a common monotonic trend or isolate a causal effect of
subsequent activity or log text volume.

\begin{table}[!htbp]
\centering
\caption{Within-source localization with DS-V4-Flash. Each method cell
reports Step Acc / Step $\pm5$ (\%). All methods share the trajectories in
each row. SWE, TB2, TP, Vita, and WA denote SWE-bench Pro, Terminal-Bench 2,
TravelPlanner, VitaBench, and WebArena Verified, respectively.}
\label{tab:within_source_distance_text}
\scriptsize
\setlength{\tabcolsep}{2.5pt}
\renewcommand{\arraystretch}{1.12}
\resizebox{\linewidth}{!}{%
\begin{tabular}{@{}llrrcccccc@{}}
\toprule
Source & Group & Actual range & $n$ & \shortstack{All-at-\\once} & \shortstack{Step-by-\\step} & \shortstack{Binary\\search} & ECHO & FALAT & RCTA \\
\midrule
\multicolumn{10}{@{}l}{\textit{Root-to-end distance (steps)}} \\
SWE & 1 & 13--137 & 43 & 0.0 / 7.0 & 0.0 / 7.0 & 0.0 / 4.7 & 9.3 / 18.6 & 2.3 / 2.3 & 53.5 / 62.8 \\
 & 2 & 138--234 & 42 & 0.0 / 2.4 & 2.4 / 11.9 & 0.0 / 2.4 & 7.1 / 19.0 & 2.4 / 9.5 & 31.0 / 38.1 \\
 & 3 & 235--605 & 43 & 4.7 / 4.7 & 0.0 / 11.6 & 2.3 / 4.7 & 7.0 / 16.3 & 2.3 / 20.9 & 30.2 / 39.5 \\
TB2 & 1 & 26--491 & 42 & 7.1 / 21.4 & 4.8 / 16.7 & 4.8 / 23.8 & 21.4 / 38.1 & 2.4 / 16.7 & 26.2 / 45.2 \\
TP & 1 & 0--16 & 227 & 4.8 / 10.1 & 4.0 / 11.9 & 0.0 / 2.2 & 7.9 / 13.2 & 1.8 / 9.3 & 16.3 / 31.7 \\
 & 2 & 17--87 & 229 & 5.2 / 15.3 & 2.2 / 9.2 & 1.7 / 10.0 & 13.1 / 23.1 & 3.1 / 10.5 & 18.8 / 27.9 \\
 & 3 & 88--470 & 229 & 3.5 / 20.1 & 4.4 / 17.9 & 4.4 / 16.2 & 5.7 / 20.1 & 3.9 / 10.5 & 21.4 / 30.1 \\
Vita & 1 & 0--17 & 38 & 7.9 / 28.9 & 2.6 / 21.1 & 2.6 / 15.8 & 5.3 / 26.3 & 0.0 / 7.9 & 13.2 / 44.7 \\
 & 2 & 18--34 & 35 & 40.0 / 57.1 & 25.7 / 42.9 & 28.6 / 54.3 & 42.9 / 60.0 & 8.6 / 48.6 & 48.6 / 60.0 \\
 & 3 & 35--97 & 35 & 11.4 / 40.0 & 8.6 / 28.6 & 5.7 / 28.6 & 20.0 / 42.9 & 2.9 / 25.7 & 14.3 / 37.1 \\
WA & 1 & 2--9 & 49 & 28.6 / 57.1 & 16.3 / 42.9 & 2.0 / 8.2 & 34.7 / 46.9 & 0.0 / 12.2 & 46.9 / 69.4 \\
 & 2 & 16--33 & 62 & 21.0 / 38.7 & 14.5 / 29.0 & 3.2 / 19.4 & 30.6 / 40.3 & 0.0 / 3.2 & 40.3 / 50.0 \\
 & 3 & 34--62 & 66 & 4.5 / 16.7 & 4.5 / 18.2 & 9.1 / 31.8 & 15.2 / 30.3 & 6.1 / 22.7 & 16.7 / 39.4 \\
\midrule
\multicolumn{10}{@{}l}{\textit{Log text volume (characters)}} \\
SWE & 1 & 117,109--316,399 & 42 & 0.0 / 2.4 & 2.4 / 9.5 & 0.0 / 0.0 & 11.9 / 26.2 & 2.4 / 9.5 & 50.0 / 61.9 \\
 & 2 & 321,686--509,229 & 43 & 0.0 / 7.0 & 0.0 / 11.6 & 0.0 / 7.0 & 4.7 / 9.3 & 2.3 / 4.7 & 44.2 / 53.5 \\
 & 3 & 510,282--1,224,895 & 43 & 4.7 / 4.7 & 0.0 / 9.3 & 2.3 / 4.7 & 7.0 / 18.6 & 2.3 / 18.6 & 20.9 / 25.6 \\
TB2 & 1 & 42,854--870,799 & 42 & 7.1 / 21.4 & 4.8 / 16.7 & 4.8 / 23.8 & 21.4 / 38.1 & 2.4 / 16.7 & 26.2 / 45.2 \\
TP & 1 & 23,125--129,713 & 228 & 10.1 / 22.8 & 5.7 / 18.0 & 2.6 / 14.5 & 13.6 / 27.6 & 3.5 / 17.1 & 21.5 / 38.2 \\
 & 2 & 129,750--185,641 & 228 & 2.2 / 14.9 & 2.2 / 11.8 & 1.8 / 7.0 & 7.9 / 15.8 & 2.6 / 8.3 & 17.5 / 25.9 \\
 & 3 & 185,763--504,578 & 229 & 1.3 / 7.9 & 2.6 / 9.2 & 1.7 / 7.0 & 5.2 / 13.1 & 2.6 / 4.8 & 17.5 / 25.8 \\
Vita & 1 & 1,231--13,740 & 36 & 38.9 / 72.2 & 25.0 / 55.6 & 27.8 / 55.6 & 41.7 / 66.7 & 5.6 / 52.8 & 41.7 / 80.6 \\
 & 2 & 14,210--23,198 & 36 & 8.3 / 25.0 & 5.6 / 19.4 & 5.6 / 25.0 & 16.7 / 33.3 & 2.8 / 19.4 & 22.2 / 30.6 \\
 & 3 & 23,519--224,950 & 36 & 11.1 / 27.8 & 5.6 / 16.7 & 2.8 / 16.7 & 8.3 / 27.8 & 2.8 / 8.3 & 11.1 / 30.6 \\
WA & 1 & 28,733--131,806 & 59 & 15.3 / 42.4 & 10.2 / 35.6 & 3.4 / 22.0 & 27.1 / 39.0 & 3.4 / 15.3 & 32.2 / 50.8 \\
 & 2 & 132,101--141,128 & 59 & 25.4 / 40.7 & 15.3 / 28.8 & 1.7 / 20.3 & 32.2 / 39.0 & 0.0 / 8.5 & 37.3 / 50.8 \\
 & 3 & 141,251--197,322 & 59 & 10.2 / 23.7 & 8.5 / 22.0 & 10.2 / 20.3 & 18.6 / 37.3 & 3.4 / 15.3 & 30.5 / 52.5 \\
\bottomrule
\end{tabular}%
}
\end{table}

\FloatBarrier

\section{Root-cause decision rules}
\label{app:rules}

The final-attribution stage applies the operational definition in
Section~\ref{sec:benchmark} to the recorded trajectory. The rules below specify
how it resolves common boundary cases and selects a single root-cause step.

\begin{itemize}
\item \textbf{Error already present in a handoff instruction.} If a later step
implements an erroneous decision already stated in an earlier handoff
instruction, the handoff step is selected as the root-cause step.
\item \textbf{Error introduced after a handoff.} If a later step departs from
the handoff instruction or introduces a decisive error not present in it, that
later step is selected as the root-cause step.
\item \textbf{Successfully repaired error.} An earlier error is excluded if it
is corrected before the execution ends.
\item \textbf{Error missed during later verification.} A later verifier may
fail to detect an earlier error. This omission may be retained as unscored
explanatory context, but it does not replace the step that introduced the
error.
\item \textbf{Contradictory task and evaluation requirements.} An input step
may be selected only if the task instruction and the official evaluator or test
requirement are explicitly incompatible. Both texts must be cited from the
recorded trajectory.
\item \textbf{Earliest supported introduction.} Among candidates supported by
the original log, RCTA selects the earliest recorded step that introduced a
decisive error relevant to the failed task execution. The selected error must
remain unrepaired when the execution ends.
\end{itemize}

These rules govern only the root-cause-step output. RCTA predicts and scores the
responsible role in a separate output field.

\section{RCTA prompts and implementation details}
\label{app:prompts}

At the implementation level, RCTA uses three prompt stages: local summarization,
trajectory-outline aggregation, and final attribution. These implement, but are
not identical to, the two scientific stages and four components in
Figure~\ref{fig:pipeline}. A \emph{segment} is a consecutive block of original
steps; a \emph{phase} groups adjacent segment summaries serving the same
subgoal without repartitioning recorded steps. A \emph{handoff instruction} is a logged \texttt{X (-> Y)} message.

For segmentation, a \emph{verifier signal} is a rule-extracted
\texttt{PASS} or \texttt{FAIL} marker from explicit language in a verifier or
terminal record. The signal serves only as an optional boundary cue, not as a
semantic judgment, benchmark label, or the source evaluator's final outcome.

RCTA applies the local-summary prompt once per segment, followed by one
trajectory-outline call and one final-attribution call. A trajectory with $m$
segments therefore requires $m+2$ initial LLM calls before any
validation-triggered retry. All 1,140 full-benchmark RCTA predictions use the
same three system-prompt templates for local summarization, trajectory-outline
aggregation, and final attribution, provided with the implementation and the
code for constructing user messages. The prompts
contain task rules and output-schema guidance, but no LongRCA Bench instances
or input--output demonstrations. User messages are constructed from the
released identifier and trajectory history. The reference role, root step, and
rationale fields are not provided during prediction.

\begin{table}[!htbp]
\centering
\caption{Prompt requirements and programmatic membership checks for the three
RCTA prompt templates. The local-summary template is applied once per segment;
the final role and root fields are validated independently.}
\label{tab:prompt_contracts}
\normalsize
\setlength{\tabcolsep}{3.5pt}
\renewcommand{\arraystretch}{1.08}
\begin{tabularx}{\linewidth}{@{}l >{\raggedright\arraybackslash}p{0.22\linewidth} >{\raggedright\arraybackslash}X >{\raggedright\arraybackslash}p{0.23\linewidth}@{}}
\toprule
\textbf{Stage} & \textbf{Input} & \textbf{Required output} & \textbf{Validation} \\
\midrule
Local & Segment, task, and overlap & Purpose and candidate steps with evidence IDs & All IDs occur in the segment \\
Outline & Segment summaries and verifier-signal summary & Contiguous subgoal phases with step IDs & All IDs occur in the trajectory \\
Attribution & Outline, summaries, candidate text, and handoffs & Independent role and root fields, evidence IDs, and any cited handoff & Role and IDs occur in the trajectory; cited text is grounded \\
\bottomrule
\end{tabularx}

\end{table}

The outline-stage validator in Table~\ref{tab:prompt_contracts}
checks referenced step identifiers. If the phase-construction response cannot
be parsed, the implementation assigns one phase to each local summary.

The final-attribution validator checks the role and root fields independently.
It accepts only a role label recorded in the trajectory and a root step drawn
from the same trajectory. An invalid field does not overwrite a valid
prediction in the other field.
External subjects such as \texttt{ENVIRONMENT} are not benchmark labels and
cannot be scored outputs. The marker \texttt{system\_evaluation} may appear only
as an unscored terminal node in the explanatory failure chain; it cannot be the
predicted root step. Every supporting step must also belong to the recorded
trajectory.

Segmentation uses rule-based implementation settings, not learned parameters.
When adding the next step would exceed 32,000 characters, the implementation
starts a new segment when possible; an individual overlong step may exceed this
limit. Each segment contains at most 80 non-overlap steps. A boundary after a
finish step, or before a role-changing handoff instruction or completion step,
requires at least eight steps and 10,000 characters in the current segment. A
boundary before a change in the most recent explicit verifier signal requires
10,000 characters but no eight-step minimum. After the 24,000-character target,
an ordinary source-role change can form a soft boundary.

Except for the first, each segment receives up to five preceding steps from the
previous segment as overlap context. The base candidate list is capped at 80
before preceding handoff instructions are added, so the final excerpt set can
exceed 80. For an upstream attribution, the checker accepts a normalized
substring match or sufficient word overlap between the cited text and the
referenced handoff instruction. This check establishes textual provenance but
does not independently verify that the cited instruction contains the same
causal error.

\paragraph{Candidate ordering and role matching.}
Candidate hints are deduplicated by integer step ID, merging distinct role,
source, and note fields. The fixed retrieval priority is possible
manifestation (0), possible root (1), terminal context (2), and other
candidates (3); a recovered-noise hint adds 3. If several tags occur, the
first matching category in this order sets the base priority. Ties use
ascending step ID. The first 80 candidates are retained before handoff
augmentation; the expanded set is deduplicated without a second 80-step
cap. Actual excerpts contain only IDs present in the trajectory. These
rules prioritize evidence retrieval, not the final root-cause decision.
They are implemented by \texttt{\_add\_candidate},
\texttt{\_candidate\_priority}, and
\texttt{\_collect\_candidate\_step\_hints} in
\texttt{analyze/global\_reducer.py} in the released implementation.

Role classes are determined by recorded names in
\texttt{preprocess/step\_enricher.py}. Planner names are
\texttt{DiagnostAgent}, \texttt{Task\_Planner}, \texttt{Orchestrator}, and
\texttt{manager\_agent}; executor names are \texttt{ActionAgent},
\texttt{Action\_Expert}, \texttt{WebSurfer}, \texttt{FileSurfer},
\texttt{Assistant}, \texttt{Coder}, \texttt{transport\_agent},
\texttt{stay\_agent}, and \texttt{food\_attr\_agent}; verifier names are
\texttt{JudgeAgent}, \texttt{Verification\_Expert}, and
\texttt{verifier\_agent}. Classification trims whitespace and extracts the
sender from a handoff name. Names ending in \texttt{Expert} or
\texttt{\_Expert} map to verifier if they contain \texttt{verif} or
\texttt{validat} (case-insensitive), and otherwise to executor.
\texttt{Computer\_terminal} and \texttt{ComputerTerminal} map to terminal,
and \texttt{human} maps to human; unmatched names remain unknown.
An \texttt{Orchestrator} status suffix retains the planner class.

For executor or verifier candidates, handoff augmentation requires the
logged recipient to equal the candidate's raw agent name, with a
case-sensitive comparison; matching the broad role class is insufficient.
It selects the nearest strictly earlier matching handoff and uses no
cross-recipient fallback. Other candidate roles use the nearest earlier
handoff regardless of recipient. The relevant functions are
\texttt{\_delegate\_target}, \texttt{\_nearest\_preceding\_delegate}, and
\texttt{\_augment\_with\_preceding\_delegates} in the same reducer module.

The implementation also contains role-name-triggered recall rules for a
configured family of planner--executor logs. These rules can add visible
executor errors, planner decisions, recovered-error context, and terminal
context before candidate prioritization. None of the 1,140 LongRCA Bench
trajectories matched the configured role-name signature, so these rules did not
affect the LongRCA results.

\paragraph{Ablation inference settings.}
Component comparisons use DS-V4-Flash with output budgets of 4{,}000 tokens
for local summaries (reasoning off), 12{,}000 for aggregation, and 32{,}000
for attribution. One length-triggered retry doubles the budget, subject to
a 65{,}536-token cap. Attribution requests temperature 0; aggregation and
attribution use reasoning at high effort. All component conditions share
these settings. Full-benchmark evaluation uses output budgets of
4{,}000, 6{,}000, and 8{,}000 tokens for local summarization, aggregation,
and attribution, respectively.

\paragraph{Component-control implementation.}
The A3 control omits explicit backward-reasoning instructions while keeping
the original candidate and handoff texts and step IDs. It removes linkage
and role hints, replaces annotations for added
handoff excerpts with neutral context labels, and omits the
instruction-evidence check. The A4 control uses Full's initial attribution
output after any length-triggered retry and omits validation-driven retry
and output sanitization. Its output is checked only to measure validity.

\paragraph{Ablation statistical protocol.}
Component comparisons use source-macro aggregation over the five sources
and four pre-specified exact McNemar tests on Step Acc against Full RCTA,
with Holm correction (family $\alpha{=}.05$).
Source-stratified paired bootstrap CIs (10{,}000 resamples; seed 20260820)
are pointwise and quantify trajectory resampling conditional on the observed
outputs. At the observed discordance rates (${\approx}0.15$), a post-hoc
sensitivity calculation estimates 80\% power for effects of roughly 9--10
percentage points in the four-test design. Accuracy uses a fixed denominator
of 200; abstained, missing, or unparseable predictions count as errors.

\paragraph{Inference-time repeatability.}
Across three pipeline realizations on a 30-trajectory panel, Full RCTA
obtains 26.7/30.0/43.3\% Step Acc, compared with 26.7/16.7/30.0\% without
A1 and 23.3/20.0/33.3\% without A2. The corresponding Full-minus-ablation
differences are 0.0/13.3/13.3 and 3.3/10.0/10.0 percentage points. Pairwise
agreement between Full root-step predictions is 50.0--56.7\%.



\paragraph{Candidate-retention diagnostic.}
The base-candidate set and the handoff-augmented excerpt set are evaluated
separately. Relative to Full, the equal-size uniform control changes
base-candidate recall by $-31.5$ pp (pointwise 95\% CI $[-39,-24]$),
all-excerpt recall by $-25.5$ pp ($[-33,-18]$), final-evidence recall
by $-13.0$ pp ($[-19,-7]$), and Step Acc by $-6.0$ pp ($[-11.0,-1.0]$).
The three recall contrasts are exploratory and are not members of the
four-comparison Step Acc test family.

\begingroup
\rightskip=0pt plus 2em
\paragraph{Handoff-root diagnostic.}
On the 41 handoff-root cases within the 80-trajectory cohort
(Appendix~\ref{app:component_diagnostics}), removing A2 or A3 lowers
Step Acc by 26.8 or 19.5 pp, respectively; the unadjusted exact McNemar
$p$-values are .0034 and .077. On the 39 non-handoff-root cases, both
Step Acc differences are zero. Among non-abstaining, parseable in-range
handoff-root predictions, late-prediction rates are 40.0\%/60.0\%/51.2\%
and mean signed errors (prediction minus reference) are +24.0/+32.4/+32.3
steps for Full/without A2/without A3. These subgroup diagnostics are
exploratory.
\par
\endgroup

\section{RCTA inference pseudocode}
\label{app:pseudocode}

The pseudocode exposes the correspondence between the operational inference
sequence and A1--A4 while retaining the implementation fallbacks described in
Section~\ref{app:prompts}. Indices $c$ and $h$ denote steps $s_c$ (a base
candidate) and $s_h$ (its retrieved handoff instruction). Handoff retrieval
(A2) only augments the evidence package. The instruction--execution comparison and independent role/root
prediction occur in \textsc{CompareAndAttribute} (A3).
After the single validation-triggered retry, valid role and root fields are
preserved independently, unsupported fields are removed, the output is flagged
for review, and the system abstains only when an essential field remains
invalid.

\begin{algorithm}[!t]
\caption{RCTA inference for a failed trajectory.}
\label{alg:rcta_inference}
\normalsize
\begin{algorithmic}[1]
\Require Failed trajectory \(\tau=(s_0,\ldots,s_{T-1})\) with a normalized, step-indexed history.
\Ensure Responsible role \(\hat{\rho}\), root step \(\hat{r}\), evidence \(E\), or abstention.
\Statex \textbf{Stage I: Failure-relevant candidate recall (A1)}
\State \(\mathcal{S} \gets \Call{EnrichAndSegment}{\tau}\)
\State \(L \gets [\Call{LocalSummary}{S_j} \mid S_j \in \mathcal{S}]\)
\State \(P \gets \Call{OrganizePhases}{L}\)
\If{\(P\) is not parseable}
    \State \(P \gets \Call{SingletonPhases}{L}\)
\EndIf
\State \(B \gets \Call{RetrieveSourceSteps}{L,P}\)
\If{\(\Call{PlannerExecutorSignature}{\tau}\) matches}
    \State \(B \gets B \cup \Call{RuleBasedCandidates}{\tau}\)
\EndIf
\State \(B \gets \Call{TopPrioritized}{B,80}\)
\State \(C \gets \Call{CandidatePackage}{B,L,P,\tau}\)
\Statex \textbf{Stage II: Handoff-aware backward attribution (A2--A3)}
\ForAll{\(c \in B\)}
    \If{\(s_c\) is emitted by an executor or verifier}
        \State \(h \gets \Call{RoleMatchedHandoff}{c,\tau}\)
    \Else
        \State \(h \gets \Call{NearestEarlierHandoff}{c,\tau}\)
    \EndIf
    \State \(C \gets \Call{AttachIfPresent}{C,h}\)
\EndFor
\State \((\hat{\rho},\hat{r},E) \gets \Call{CompareAndAttribute}{P,C}\)
\Statex \textbf{Reliability layer: Validation (A4)}
\State \(v \gets \Call{ValidateFields}{\hat{\rho},\hat{r},E,\tau}\)
\If{\(v\) reports an invalid field}
    \State \((\hat{\rho},\hat{r},E) \gets \Call{RetryWithFeedback}{P,C,v}\)
    \State \(v \gets \Call{ValidateFields}{\hat{\rho},\hat{r},E,\tau}\)
\EndIf
\State \Return \(\Call{SanitizeOrAbstain}{\hat{\rho},\hat{r},E,v}\)
\end{algorithmic}
\end{algorithm}

\begingroup
\let\FloatBarrier\relax
\section{Extended related work and comparison notes}
\label{app:related_work}

Broader failure-characterization resources describe errors at levels that
complement agent and step attribution. MAST organizes recurring failure
modes, whereas TRAIL attaches error types and impact to execution spans
\citep{mast,trail}. MP-Bench accommodates multiple plausible failure-inducing
steps \citep{mpbench}. Among the resources in
Table~\ref{tab:benchmark_landscape}, Who\&When Pro scales data collection
through controlled error injection, while TraceElephant emphasizes complete
execution observability \citep{whowhenpro,traceelephant}.

RAFFLES uses iterative judge--evaluator feedback, whereas AgentRx grounds
diagnosis in checked constraint violations \citep{raffles,agentrx}. CDC
combines Shapley-based agent attribution with temporal causal discovery for
critical-step prediction \citep{cdc}. Diagnostic support can also extend
beyond the given trajectory. FAMAS re-executes failed tasks and ranks actions
using spectrum statistics across runs, while CORRECT retrieves error schemata
distilled from previously annotated failures \citep{famas,correct}.

In Table~\ref{tab:benchmark_landscape}, \emph{Error Source} distinguishes
failures arising during ordinary task execution (Natural) from failures
constructed through deliberate error injection (Synthetic). Controlled task
conditions, such as HORIZON's horizon settings, are distinct from deliberate
error injection. \emph{Diagnostic Targets} identifies the annotated object:
Agent denotes the responsible agent or workflow role, Component an execution
component, and Type a failure-mode or error-cause category. Step and Steps
retain the localization targets summarized in the table's superscript notes;
their annotation rules remain benchmark-specific.

\emph{Total (Reviewed)} reports the number of released or analyzed
trajectories, followed by the human-reviewed count in parentheses. Equal
counts indicate review of all counted trajectories, but do not imply identical
review or annotation protocols across resources. \emph{Avg. length (steps)}
reports each resource's overall mean using its own step-counting convention;
-- denotes an unreported mean. For Failure as a Process, both trajectory
counts and average length refer to failed runs.

\endgroup

\section{Benchmark construction and supplementary details}
\label{app:benchmark_details}

\subsection{Human annotation and quality control}
\label{app:annotation}

\begin{figure}[!t]
\centering
\includegraphics[width=\linewidth]{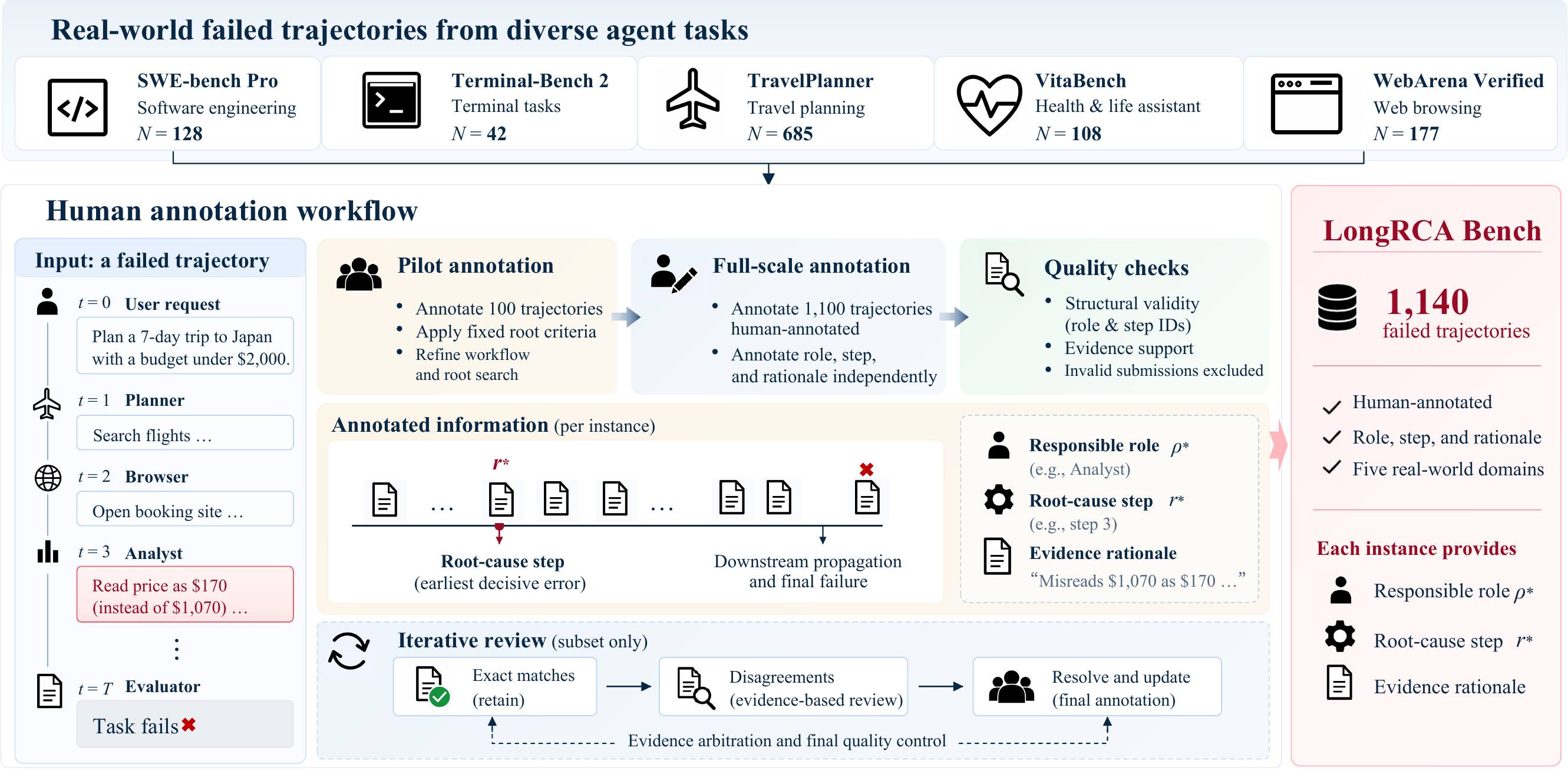}
\caption{Construction and annotation workflow for LongRCA Bench. Source runs
are filtered and normalized into step-indexed histories before annotation,
label resolution, and final quality control.}
\label{fig:annotation_pipeline}
\end{figure}

The same annotator pool participated in calibration and full-scale annotation.
Calibration refined how they navigated long histories, traced candidate
errors, and documented evidence; the root-cause definition and decision rules
remained unchanged. In both stages, annotators independently reviewed task
instructions, complete step-indexed histories, and source evaluator outcomes,
then submitted a responsible role, root step, and evidence-grounded rationale.

DeepSeek V4 Flash Preview generated factual activity summaries to support
navigation. The prompt requested descriptions of recorded activities without
root-cause judgments. Summaries remained separate from annotation fields;
reference labels and rationales required support from task requirements and
original trajectory records.

Full-scale submissions with verifiable role, step, and rationale fields
number 1,343 over 1,041 trajectories: 747 have one annotation, 286 have two,
and eight have three. The multiply annotated subset therefore contains 294
trajectories and $286+8\binom{3}{2}=310$ unordered pairs, each from different
annotators. Pairwise agreement is computed before label resolution, with
role names normalized as in Appendix~\ref{app:scoring_protocols}. Exact role,
step, and joint matches number 205, 125, and 121 (66.1\%, 40.3\%, and 39.0\%).
Calibration is reported separately in Table~\ref{tab:annotation_summary};
the two sets are not pooled. Typical annotation time was 30--40 minutes.

\begin{table}[!t]
\centering
\caption{Pre-resolution disagreement among completed, structurally valid
annotations. Each unordered pair of different annotators has equal weight.
Role agreement uses the evaluation normalization. Step-gap percentages sum
to 100 within each row; quartiles use linear interpolation. Calibration and
full-scale annotations are summarized separately.}
\label{tab:annotation_summary}
\small
\setlength{\tabcolsep}{3.5pt}
\renewcommand{\arraystretch}{1.12}
\begin{tabular}{@{}lrrrrrrrr@{}}
\toprule
Role relation & Pairs & Role & Step $\pm5$ & Median [Q1, Q3] & 0 & 1--5 & 6--20 & $>20$ \\
\midrule
\multicolumn{9}{@{}l}{\textit{Full-scale: 294 trajectories}} \\
All pairs & 310 & 66.1 & 50.3 & 5 [0, 46] & 40.3 & 10.0 & 11.6 & 38.1 \\
Same role & 205 & 100.0 & 67.3 & 0 [0, 15] & 59.0 & 8.3 & 8.3 & 24.4 \\
Different role & 105 & 0.0 & 17.1 & 42 [13, 78] & 3.8 & 13.3 & 18.1 & 64.8 \\
\midrule
\multicolumn{9}{@{}l}{\textit{Calibration: 100 trajectories}} \\
All pairs & 296 & 53.7 & 37.8 & 17.5 [1, 54.25] & 24.7 & 13.2 & 15.5 & 46.6 \\
Same role & 159 & 100.0 & 59.7 & 1 [0, 29] & 45.9 & 13.8 & 11.9 & 28.3 \\
Different role & 137 & 0.0 & 12.4 & 37 [15, 82] & 0.0 & 12.4 & 19.7 & 67.9 \\
\bottomrule
\end{tabular}
\end{table}

The full-scale step gap has median 5 and interquartile range 0--46.
Only 31 of the 185 non-exact pairs (16.8\%) differ by 1--5 steps; nearby
localization therefore explains a minority of exact-step disagreements.
The median gap is 0 for same-role pairs and 42 for different-role pairs.
These descriptive results concern the existing multiply annotated samples.

A valid submission must select an integer step within the trajectory range
and a role present in the recorded role set. Its rationale must identify
trajectory evidence supporting the proposed causal interpretation.
Submissions with invalid references, evident annotation defects, or
insufficient evidence are excluded; selected steps are not automatically
changed to repair submissions.

For trajectories with multiple valid annotations, labels are compared field
by field, and exact matches are retained as consensus references. Two experts
resolve disagreements using the original task, complete history, source
evaluator outcome, and written rationales under the attribution criteria.
Final quality control re-examines every retained trajectory, including
calibration cases, and excludes unresolved attribution ambiguity, invalid
labels, or inadequate evidential support. Each released reference retains
separate role and root-step fields with a supporting rationale.

\paragraph{Sensitivity to observed root-step labels.}
We assess root-step sensitivity on the 294 multiply annotated full-scale
trajectories with predictions from all six methods. We first score against
the final reference roots, then repeat scoring over 10,000 draws (seed
20260918). Each draw uniformly selects one complete observed human
annotation per trajectory and uses its root step for all six methods.
Trajectories and predictions remain fixed, and scores weight trajectories
equally.

\begin{table}[!t]
\centering
\caption{Root-step sensitivity to observed human labels on 294 full-scale
trajectories with DS-V4-Flash.
Accuracy cells show final-label score $\to$ resampled mean [2.5th, 97.5th
percentiles] (\%). Gap rows use RCTA minus the strongest baseline in each
draw (pp). Rank-1 rates require a strict lead; ties do not count as wins.
Intervals describe label sensitivity on fixed samples, not population
confidence intervals.}
\label{tab:annotation_sensitivity}
\small
\setlength{\tabcolsep}{3.5pt}
\renewcommand{\arraystretch}{1.12}
\begin{tabular}{@{}lcc@{}}
\toprule
Method / comparison & Step Acc & Step $\pm5$ \\
\midrule
All-at-once & 7.8 $\to$ 8.3 [7.1, 9.5] & 21.1 $\to$ 22.1 [19.7, 24.5] \\
Step-by-step & 6.1 $\to$ 6.5 [5.4, 7.5] & 15.6 $\to$ 17.3 [15.3, 19.4] \\
Binary search & 3.7 $\to$ 3.9 [3.1, 4.8] & 15.0 $\to$ 15.9 [13.9, 18.0] \\
ECHO & 17.0 $\to$ 17.6 [15.6, 19.4] & 27.9 $\to$ 31.1 [28.6, 33.7] \\
FALAT & 1.0 $\to$ 2.2 [1.4, 3.1] & 13.6 $\to$ 14.4 [12.2, 16.7] \\
RCTA & 21.8 $\to$ 23.5 [21.1, 25.9] & 35.4 $\to$ 36.2 [33.7, 38.8] \\
\addlinespace[2pt]
RCTA gap (pp) & 4.8 $\to$ 5.9 [3.1, 8.8] & 7.5 $\to$ 5.1 [1.4, 8.5] \\
RCTA strict rank 1 & 100.00\% & 99.68\% \\
Full ordering retained & 98.39\% & 72.15\% \\
\bottomrule
\end{tabular}
\end{table}

On the 294-trajectory subset, RCTA strictly leads all baselines in 100.00\%
of draws for Step Acc and 99.68\% for Step $\pm5$; its gap to the strongest
baseline has a central 95\% range of [3.1, 8.8] and [1.4, 8.5] pp,
respectively. Full method rankings vary more than the leading step method
(Table~\ref{tab:annotation_sensitivity}). These results describe root-step
sensitivity within the available annotation overlap.

\paragraph{Automatic release checks.}
All 1,140 released records have an existing integer root index, a role in the
recorded role set, a nonempty rationale, unique contiguous step indices,
matching individual and merged records, and a verified mapping to source
trajectory content. These checks
establish structural integrity and traceability, not causal correctness;
neither agreement nor label sensitivity estimates final-label accuracy or
independent expert reproducibility \citep{klie2024annotationquality}.

\paragraph{Responsible role and the recorded event.}
The role and root step are recorded separately, with a rationale grounded
in the task and trajectory. A log event can relay another agent's action
or completion report. For example, in \texttt{swe\_bench\_pro\_\_047},
step 133 is logged under \texttt{Computer\_terminal} but relays
\texttt{ActionAgent}'s delegated completion report. The retained human
reference assigns responsibility to \texttt{ActionAgent}, with step 133
as the root. Its rationale identifies the missing required configuration
file despite the success report, and distinguishes later missed checks
from that failure. This example shows why the event's outer sender does
not mechanically determine the reference role.

\subsection{Complete illustrative case}
\label{app:case_study}

Figure~\ref{fig:case0104} shows a pricing error in a 223-step TravelPlanner
trajectory.\footnote{Released as instance \texttt{travelplanner\_\_021} in
LongRCA Bench, corresponding to TravelPlanner test query 104, generated by
Kimi-K2.5 with Magentic-One group-chat coordination (run 0001).}
The user requests a three-day trip for one traveler from Plattsburgh to Fort
Lauderdale within a \$1,200 budget. The human reference assigns responsibility
to \texttt{stay\_agent} and identifies step 73 as the root-cause step.

The accommodation-search result at step 72 contains prices for multiple
properties. In one subsequent output, \texttt{stay\_agent} issues five
candidate-recording calls, represented as steps 73--77 in the normalized
history. Step 73 records Clean 1br at ``\$483 for 2 nights,'' and step 74
records Lefferts Garden 1 bedroom at ``\$317 for 2 nights.'' The remaining
calls likewise interpret candidate prices as two-night totals. The reference
step is the first candidate record in this output, while the displayed
budget path follows Lefferts, the property ultimately selected. The calls
document repeated misinterpretation within one output; the log does not
establish a direct dependency from the step-73 candidate record to step 74.

Steps 78 and 79 confirm that both candidate notes were stored on the
blackboard. At step 83, the agent compares the candidates and recommends
Lefferts, calculating a \$172 remainder using its \$317 lodging total.
The planner allocates this remainder to food and activities at step 86.
The workflow verifier approves the budget and later rechecks the accommodation
after transport changes, retaining the same two-night interpretation. During
finalization, the manager corrects a separate arithmetic error in the
food/activity allocation from \$357 to \$257, but leaves the lodging total
at \$317. The correction therefore repairs the allocation arithmetic while
preserving the erroneous price interpretation.

The final itinerary at step 220 lists Lefferts for both nights without
stating a lodging price. TravelPlanner defines \texttt{price} as a nightly
rate, but the original accommodation-search response does not specify the
unit. Applying this pricing rule to the
recorded itinerary gives flight costs of \(\$341+\$255=\$596\) and lodging
costs of \(\$317\times2=\$634\). These items alone total \$1,230, exceeding
the budget before food or local transport. This budget check is part of
retrospective diagnosis; the recorded source evaluator reports commonsense
failures without executing the budget test. The case shows how an early
misinterpretation can persist through verification and local correction,
requiring the final plan to be read together with intermediate records and
the applicable pricing rule.

\subsection{Additional benchmark statistics}
\label{app:benchmark_statistics}

Median trajectory length ranges from 51 recorded steps for VitaBench to 257.5
for SWE-bench Pro; source medians for root-to-end distance range from 24 to
186 steps. Figure~\ref{fig:delay} reports the overall distribution summaries
and threshold counts. Table~\ref{tab:mini_structure} retains the total recorded
step count, and Table~\ref{tab:mini_source_structure} reports source-level
means alongside the Mini comparison.

\section{Supplementary evaluation protocols and diagnostics}
\label{app:evaluation_details}

\subsection{Scoring and experiment protocols}
\label{app:scoring_protocols}

For the benchmark comparisons, role names are normalized for case,
whitespace, and explicit handoff suffixes. Agent and step accuracies are
computed from their respective output fields. Every expected trajectory
remains in the denominators; invalid fields count as errors for their
respective targets.

All component conditions use the same reference labels and scoring rules.
Role matching normalizes case and whitespace while retaining handoff suffixes;
abstentions count as errors.

Component coverage is the percentage of trajectories with non-abstaining
outputs that pass the validation checks for their condition. These checks cover recorded
role names, root-step validity, explanatory-chain structure, and cited
handoff evidence. Appendix~\ref{app:prompts} specifies the A3 control's changes
to these checks and the A4 control's use of validity checks only for scoring.

For the full-benchmark comparison, we also report \textbf{source-weighted valid-output root
MAE}. Let $\mathcal{V}_b$ be the trajectories in source $b$ with a valid,
in-range numeric root-step prediction, $N_b$ the total number of trajectories from that
source, and $N=\sum_b N_b$. The metric is
\[
\frac{1}{N}\sum_b N_b
\left(\frac{1}{|\mathcal{V}_b|}\sum_{i\in\mathcal{V}_b}
|\hat{r}_i-r_i^*|\right).
\]
We average absolute root-step errors over valid predictions within each
source, then weight these source means by the original source proportions.
Missing or invalid step outputs count as errors for both root-step accuracy
metrics but are excluded from the source means used for MAE. Exact matching
remains the primary localization metric. A fixed step tolerance represents
different fractions of trajectories whose lengths differ by an order of
magnitude.

Fidelity evaluates the fixed Mini QIDs using the same DS-V4-Flash predictions
as the full benchmark.
Appendix~\ref{app:prompts} records prompt requirements and validation rules;
Algorithm~\ref{alg:rcta_inference} gives the field-preserving fallback sequence.

\subsection{Full-benchmark supplementary metrics}
\label{app:full_metrics}

\begin{table}[!t]
\centering
\caption{Full-benchmark results on all 1,140 trajectories. All methods use
DS-V4-Flash. Agent Acc measures agreement with the reference responsible role. Root MAE is the source-weighted valid-output mean
absolute step error. Higher values are better for the three accuracy metrics,
whereas lower values are better for MAE.}
\label{tab:main_results_full}
\normalsize
\fontencoding{OT1}\selectfont
\setlength{\tabcolsep}{5pt}
\renewcommand{\arraystretch}{1.12}
\begin{tabular}{@{}lrrrr@{}}
\toprule
\textbf{Method} & \textbf{Agent Acc$\uparrow$} &
\textbf{Step Acc$\uparrow$} & \textbf{Step $\pm5$$\uparrow$} &
\textbf{Root MAE$\downarrow$} \\
\midrule
All-at-once & 26.2\% & 7.6\% & 19.9\% & 55.9 \\
Step-by-step & 22.2\% & 5.3\% & 16.9\% & 52.3 \\
Binary search & 23.0\% & 3.4\% & 13.3\% & 61.7 \\
ECHO & 27.5\% & 13.2\% & 24.7\% & 50.4 \\
FALAT & 19.0\% & 2.8\% & 12.5\% & 66.6 \\
\textbf{RCTA (ours)} & \textbf{51.1\%} & \textbf{24.1\%} &
\textbf{37.4\%} & \textbf{38.6} \\
\bottomrule
\end{tabular}

\end{table}

RCTA, as shown in Table~\ref{tab:main_results_full}, achieves 51.1\% Agent Acc,
24.1\% Step Acc, and 37.4\% Step $\pm5$, with a source-weighted valid-output
root MAE of 38.6 steps. Relative to ECHO, the strongest evaluated baseline on
all four metrics, its accuracy gains computed from unrounded values are
23.6, 11.0, and 12.6 percentage points, respectively, and its MAE decreases
by 11.8 steps. Despite these gains, the exact reference root is recovered
in fewer than one quarter of the trajectories.

\paragraph{Role priors.}
The majority-role baseline always predicts the most frequent reference role
within each source, using frequencies from the evaluated set
(Table~\ref{tab:role_priors}). RCTA matches this baseline on VitaBench and
exceeds it on the other four sources.

\begin{table}[!htbp]
\centering
\caption{Reference-role distributions and majority-role baselines on Full
($N=1{,}140$). Role counts are listed in descending order. $K$ is the number
of distinct normalized roles in each history, reported as median [range]
when variable. Majority and RCTA columns report Agent Acc (\%); Overall
weights trajectories equally. Source abbreviations follow
Table~\ref{tab:main_results}.}
\label{tab:role_priors}
\small
\setlength{\tabcolsep}{3pt}
\renewcommand{\arraystretch}{1.10}
\begin{tabularx}{\linewidth}{@{}lc>{\raggedright\arraybackslash}Xrr@{}}
\toprule
\textbf{Source} & \textbf{$K$} & \textbf{Reference roles (counts)} &
\textbf{Majority} & \textbf{RCTA} \\
\midrule
SWE & 5 & \texttt{DiagnostAgent} (79), \texttt{ActionAgent} (41),
\texttt{JudgeAgent} (4), \texttt{human} (3), \texttt{Computer\_terminal} (1) & 61.7 & 63.3 \\
TB2 & 5 & \texttt{DiagnostAgent} (20), \texttt{ActionAgent} (18),
\texttt{JudgeAgent} (2), \texttt{Computer\_terminal} (1), \texttt{human} (1) & 47.6 & 57.1 \\
TP & 6 [3--7] & \texttt{stay\_agent} (174), \texttt{manager\_agent} (162),
\texttt{Task\_Planner} (103), \texttt{transport\_agent} (84),
\texttt{verifier\_agent} (67), \texttt{food\_attr\_agent} (56),
\texttt{Computer\_terminal} (39) & 25.4 & 41.2 \\
Vita & 2 & \texttt{Action\_Expert} (89), \texttt{Computer\_terminal} (19) & 82.4 & 82.4 \\
WA & 4 & \texttt{Action\_Expert} (97), \texttt{Task\_Planner} (43),
\texttt{Verification\_Expert} (29), \texttt{Computer\_terminal} (8) & 54.8 & 60.5 \\
\midrule
Overall & -- & -- & 40.3 & 51.1 \\
\bottomrule
\end{tabularx}
\end{table}

\subsection{Component analysis and diagnostics}
\label{app:component_diagnostics}

\begin{table}[!t]
\centering
\caption{Component ablations on the fixed 200-trajectory sample. Rates are
source-macro percentages. MAE uses non-abstaining outputs with parseable in-range steps
($n{=}196/188/196/197/192$); Cov. uses all trajectories as its denominator and requires
non-abstaining outputs that pass the condition's validation checks (Appendix~\ref{app:scoring_protocols}).
$\Delta$ is ablation Step Acc minus that of RCTA (all components). Intervals are pointwise;
$p_{\mathrm{Holm}}$ controls the four-test Step Acc family.}
\label{tab:ablation_full}
\normalsize
\fontencoding{OT1}\selectfont
\setlength{\tabcolsep}{2.2pt}
\renewcommand{\arraystretch}{1.05}
\begin{tabularx}{\linewidth}{@{}>{\raggedright\arraybackslash}Xrrrrrrr@{}}
\toprule
Setting & \shortstack{Agent\\Acc$\uparrow$} & \shortstack{Step\\Acc$\uparrow$} &
\shortstack{Step\\$\pm5\,\uparrow$} & MAE$\downarrow$ & Cov.$\uparrow$ &
$\Delta$ [95\% CI] & $p_{\mathrm{Holm}}$ \\
\midrule
\textbf{RCTA (all components)} & 56.0 & 28.0 & 41.0 & 28.3 & 97.0 & -- & -- \\
w/o candidate-information package (A1) & 53.5 & 22.0 & 39.0 & 30.5 & 91.5 & $-6.0$ [$-11.0$, $-1.0$] & .17 \\
w/o handoff evidence (A2) & 53.5 & 24.5 & 38.0 & 29.7 & 97.0 & $-3.5$ [$-9.0$, $+2.0$] & .75 \\
w/o backward attribution (A3) & 51.0 & 25.0 & 41.0 & 31.4 & 97.5 & $-3.0$ [$-8.5$, $+2.5$] & .75 \\
w/o validation (A4) & 54.0 & 26.5 & 39.5 & 29.2 & 89.0 & $-1.5$ [$-3.5$, $0.0$] & .75 \\
\bottomrule
\end{tabularx}
\end{table}

Table~\ref{tab:ablation_full} retains the complete metrics and pointwise
intervals, which capture trajectory resampling and exclude inference-time
variability. RCTA (all components) denotes the complete component configuration on the
fixed sample (40 trajectories per source; sampling seed 20260821), evaluated
under the ablation inference settings in Appendix~\ref{app:prompts}.

\paragraph{A1: Root-evidence recall.}
The candidate-information package retains more reference-root evidence
than the equal-size uniform control (Figure~\ref{fig:ablation_mechanism}(a)).
Base-candidate recall is 19.5\% for the control and 51.0\% for Full;
recall in the handoff-augmented excerpt set is 37.5\% and 63.0\%, respectively.
The difference narrows at final evidence (25.5\% versus 38.5\%) and
Step Acc (22.0\% versus 28.0\%). This contrast evaluates excerpt
selection and the accompanying location and provenance cues as a single
package.

\paragraph{A2--A3: Handoff evidence and backward attribution.}
The diagnostic includes all 80 trajectories in the component sample with an explicit logged
handoff: 40 from SWE-bench Pro and 40 from Terminal-Bench 2. The reference
root, independently of model predictions, defines 41 handoff-root cases
(25 SWE-bench Pro, 16 Terminal-Bench 2) and 39 non-handoff-root cases
(15 and 24, respectively). Both groups contain handoffs; the other 120
trajectories are outside this diagnostic.

Full RCTA and the A2 and A3 ablations obtain 51.2\%, 24.4\%, and 31.7\%
Step Acc on handoff roots, respectively, and 28.2\% each on non-handoff
roots (Figure~\ref{fig:ablation_mechanism}(b)). Removing A2 omits the added
handoff evidence; removing A3 retains the original excerpts while removing
the backward-attribution package and yields more late predictions on
handoff roots. The observed accuracy differences are concentrated in the
handoff-root group within the shared diagnosis--execution--verification
workflow. Figure~\ref{fig:ablation_mechanism} summarizes these exploratory
contrasts; Appendix~\ref{app:scoring_protocols} specifies the scoring rules.

\paragraph{A4: Output validity.}
Validation changes the final prediction object on 18 trajectories and the
parsed root on 9; exact-root correctness changes on 3 of 200, all favoring
Full. Valid-output coverage increases from 89\% to 97\%.

Appendix~\ref{app:prompts} reports the statistical protocol and the
three-realization repeatability panel.